\documentclass{article}

\usepackage{microtype}

\usepackage{amsmath,amsfonts}
\usepackage{algorithm}
\usepackage{algorithmic}
\usepackage{graphicx}
\usepackage{textcomp}
\usepackage{xcolor}
\usepackage{multirow}
\usepackage{mathtools}
\usepackage{tcolorbox}
\usepackage{capt-of}
\usepackage{enumitem}
\usepackage{makecell}
\usepackage{float}
\usepackage{fancyhdr}
\usepackage{colortbl}
\usepackage{arydshln}
\usepackage{soul}
\usepackage{wrapfig}
\usepackage{titletoc}
\usepackage{pifont}
\usepackage{tikz}
\usepackage{booktabs}
\usepackage{listings}
\usepackage{amsthm}

\definecolor{codegreen}{rgb}{0,0.6,0}
\definecolor{codegray}{rgb}{0.5,0.5,0.5}
\definecolor{codepurple}{rgb}{0.58,0,0.82}
\definecolor{backcolour}{rgb}{0.95,0.95,0.92}

\definecolor{tabgray}{gray}{0.88}
\definecolor{tabgreen}{HTML}{D9EAD3}

\definecolor{ablquant}{HTML}{1F5FAE}   % blue (quantization: algorithm + format)
\definecolor{ablpolicy}{HTML}{1F6B1F}  % green (algorithm: runtime policy)
\definecolor{ablsys}{HTML}{8A5A00}     % dark amber (GPU system / kernels)
\definecolor{ablhw}{HTML}{6B2E8F}      % purple (accelerator)
\newcommand{\abltag}[2]{\textcolor{abl#1}{#2}}

\lstdefinestyle{mystyle}{
    backgroundcolor=\color{backcolour},
    commentstyle=\color{codegreen},
    keywordstyle=\color{magenta},
    stringstyle=\color{codepurple},
    basicstyle=\ttfamily\footnotesize,
    breakatwhitespace=false,
    breaklines=true,
    captionpos=b,
    keepspaces=true,
    showspaces=false,
    showstringspaces=false,
    showtabs=false,
    tabsize=2
}

\theoremstyle{remark}

\usepackage[utf8]{inputenc}

\DeclareUnicodeCharacter{2460}{\textcircled{1}}
\DeclareUnicodeCharacter{2461}{\textcircled{2}}

\definecolor{Reasoning}{HTML}{8dcaf6}
\definecolor{Transition}{HTML}{f8827e}
\definecolor{Execution}{HTML}{fbc08a}

\DeclareUnicodeCharacter{221A}{\ensuremath{\sqrt{}}}
\DeclareUnicodeCharacter{00B0}{\ensuremath{^\circ}}
\DeclareUnicodeCharacter{2264}{\ensuremath{\leq}}
\DeclareUnicodeCharacter{2265}{\ensuremath{\geq}}
\DeclareUnicodeCharacter{2248}{\ensuremath{\approx}}
\DeclareUnicodeCharacter{2260}{\ensuremath{\neq}}
\DeclareUnicodeCharacter{2019}{'}

\definecolor{violet9933FF}{HTML}{9933FF}
\definecolor{green00FF91}{HTML}{00FF91}
\definecolor{blue3333FF}{HTML}{3333FF}

\usepackage{graphicx}
\usepackage{subfigure}
\usepackage{enumitem}
\usepackage{booktabs} % for professional tables

\usepackage{hyperref}

\usepackage{cleveref}
\crefformat{section}{\S#2#1#3} % see manual of cleveref, section 8.2.1
\crefformat{subsection}{\S#2#1#3}
\crefformat{subsubsection}{\S#2#1#3}

\usepackage{xcolor}

\usepackage[accepted]{mlsys2025}

\mlsystitlerunning{AutoEMatic: Runtime-Adaptive Mixed-Precision LLM Inference via Automatic Exponent–Mantissa Refinement}

\begin{document}

\twocolumn[
\mlsystitle{Auto\underline{EM}atic: Runtime-Adaptive Mixed-Precision LLM Inference via Automatic \underline{E}xponent–\underline{M}antissa Refinement}

% It is OKAY to include author information, even for blind
% submissions: the style file will automatically remove it for you
% unless you've provided the [accepted] option to the mlsys2025
% package.

% List of affiliations: The first argument should be a (short)
% identifier you will use later to specify author affiliations
% Academic affiliations should list Department, University, City, Region, Country
% Industry affiliations should list Company, City, Region, Country

% You can specify symbols, otherwise they are numbered in order.
% Ideally, you should not use this facility. Affiliations will be numbered
% in order of appearance and this is the preferred way.
\mlsyssetsymbol{equal}{*}

\begin{mlsysauthorlist}
\mlsysauthor{Akshat Ramachandran$^\dagger$}{gt}
\mlsysauthor{Reena Elangovan}{nv}
\mlsysauthor{Rangharajan Venkatesan}{nv}
\mlsysauthor{Charbel Sakr}{nv}
\mlsysauthor{Tijmen Blankevoort}{nv}
\mlsysauthor{Brucek Khailany}{nv}
\mlsysauthor{Tushar Krishna}{gt}
\end{mlsysauthorlist}

\mlsysaffiliation{gt}{Georgia Institute of Technology}
\mlsysaffiliation{nv}{NVIDIA}

\mlsyscorrespondingauthor{Akshat Ramachandran}{akshat.r@gatech.edu}
% You may provide any keywords that you
% find helpful for describing your paper; these are used to populate
% the "keywords" metadata in the PDF but will not be shown in the document
\mlsyskeywords{}

\vskip 0.3in

%%%% Working on it
\begin{abstract}
Existing LLM post-training quantization techniques determine precision before inference begins, yet the precision a model requires varies dynamically across prompts, layers, and inference phases. We present AutoEMatic, a full-stack co-design that adapts the precision of weights, activations, and the KV cache at runtime. A lightweight policy selects each layer's precision at every prefill and decode step from two runtime signals: the entropy of the next-token distribution and the magnitude of each layer's update to its hidden states. This runtime adaptation is made practical by AutoFP, a multi-scale floating-point format that derives 2-, 4-, and 6-bit precisions from a single NVFP4 base format. We realize AutoEMatic on two targets: GPU kernels that execute AutoFP on NVFP4 Tensor Cores, and an accelerator whose vector MAC datapath executes all AutoFP formats natively. Across reasoning and agentic benchmarks, AutoEMatic improves accuracy by up to $21.7\%$ over NVFP4 and matches group-scaled FP6 accuracy on most benchmarks, using $4.9$ bits per weight versus $4.5$ for NVFP4. On GB300 GPUs, AutoEMatic retains approximately $90$–$98\%$ of NVFP4 decode throughput across the evaluated configurations. Furthermore, the AutoEMatic accelerator reduces latency by $1.6$--$2.3\times$ and energy by $7$--$23\%$ compared to an accelerator with NVFP4-only datapath at iso-accuracy.
\end{abstract}
]

% this must go after the closing bracket ] following \twocolumn[ ...

% This command actually creates the footnote in the first column
% listing the affiliations and the copyright notice.
% The command takes one argument, which is text to display at the start of the footnote.
% The \mlsysEqualContribution command is standard text for equal contribution.
% Remove it (just {}) if you do not need this facility.

%\printAffiliationsAndNotice{}  % leave blank if no need to mention equal contribution
\printAffiliationsAndNotice{$^\dagger$Work done during an internship at NVIDIA Research.} % otherwise use the standard text.

\section{Introduction}
\label{sec:introduction}
 
AI, driven by LLMs, is reaching an inflection point as the focus shifts from training to efficient inference at scale. Rising request volumes and long-horizon agentic workloads \cite{kariyappa2026sidequest} strain inference infrastructure in compute, memory bandwidth, and energy. Post-training quantization (PTQ) techniques improve inference efficiency by lowering the precision of weights, activations, or KV cache. Existing PTQ techniques \cite{lu2026mix,nvidia2025modelopt}, however, employ static precision assignments, determined before the start of inference rather than adapting to the model's evolving numerical demands during generation. These numerical demands vary dynamically with the prompt, between prefill and decode, and from one decoding step to the next. Therefore, a static assignment over-provisions precision where it is unnecessary and vice-versa. This motivates \textbf{runtime precision adaptation}: {dynamically adjusting precision across prompts, layers and inference phases to match a model's numerical demands}.

\begin{figure}[t]
    \centering
    \vspace{-3mm}
    \includegraphics[width=0.75\linewidth, keepaspectratio]{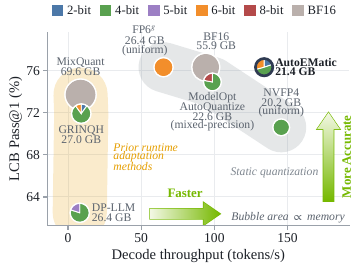}
        \vspace{-6mm}
        \caption{\textbf{LiveCodeBench accuracy versus decode throughput for Qwen3.8-27B} (batch size 1, one GB300 GPU).}
        \label{fig:teaser}
        \vspace{-7.5mm}
\end{figure}

\begin{figure*}[t]
    \centering
    \includegraphics[width=0.9\linewidth, keepaspectratio]{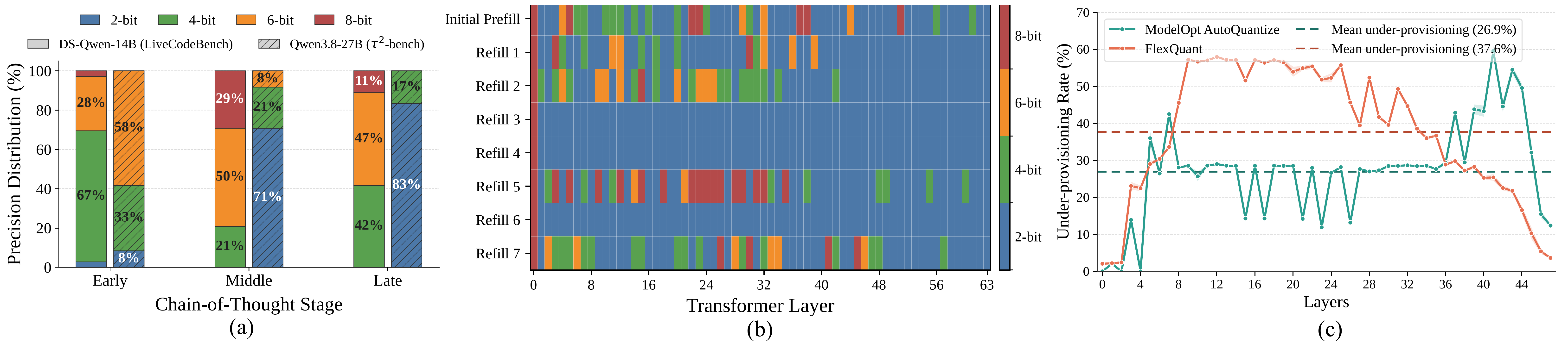}
    \vspace{-5mm}
    \caption{
    % \textbf{Precision demand shifts across decode stages, refills, and layers.}
    \textbf{(a)} Precision distribution averaged across layers in the early, middle, and late generation stages (DS-Qwen-14B, LCB; Qwen3.8-27B, $\tau^2$-bench).
    \textbf{(b)} Layer-wise precision assignments for initial prefill and additional refills (Qwen3.8-27B, $\tau^2$-bench).
    \textbf{(c)} Rate of underprovisioning for ModelOpt AutoQuantize and FlexQuant (DS-Qwen-14B, LCB).
    }
    \label{fig:motivation_1}
    \vspace{-5mm}
\end{figure*}

Recent works have taken initial steps toward this goal. FlexQuant \cite{liu2025flexquant} switches layer-wise weight precision during decoding but stores separate quantized weight copies. Multi-scale quantization \cite{nair2025matryoshka} avoids these copies by nesting several precisions within one integer bit layout whose width equals the highest supported precision. From this nested layout, lower precisions are obtained by truncating the least significant bits. DP-LLM \cite{kwon2026dp} extends this, switching each layer at decode time between two offline-calibrated precisions according to an input-dependent estimate of quantization error.

These approaches leave three challenges. First, their policies are not true runtime adaptations as they rely on offline calibration. This limits their ability to accurately track evolving precision demand during inference (\autoref{fig:teaser} (amber region)), yielding a suboptimal accuracy–precision tradeoff. Second, existing multi-scale formats use integer representations that suffer substantial accuracy degradation when lower precisions (sub-6-bit) are extracted from the nested representations, especially for activations and the KV cache (\cref{sec:results}). Their mechanism does not directly extend to group-scaled floating-point formats such as NVFP4, which is used for low-precision inference on NVIDIA Tensor Cores. Moreover, extracting precisions from their nested representations incur irregular memory accesses and dequantization overheads \cite{park2026anybcq}, that sacrifice inference performance. Finally, they do not address batched inference, where one layer must serve requests of different precisions.

\noindent
\textbf{Contributions. }To address these limitations, we introduce \textbf{AutoEMatic}. At its core is a format-agnostic runtime policy that selects each layer's precision across prefill and decode; AutoFP makes the policy practical by deriving all precisions from one NVFP4 base, and GPU kernels and an accelerator demonstrate AutoEMatic on existing and custom hardware. These four components form a full-stack co-design:

\begin{itemize}[leftmargin=*, labelsep=0.3em, itemsep=1pt, topsep=2pt, parsep=0pt, partopsep=0pt]
\item \textit{Runtime precision adaptation policy} (\cref{sec:runtime_precision_adaptation}): We identify two model-internal signals, a per-token and a per-layer measure of precision demand, and derive from their behavior a set of rules that assign layer-wise precisions.
  
\item \textit{Quantization format} (\cref{sec:quantization_format}): We introduce \emph{AutoFP}, the first multi-scale floating-point (FP) format, which composes 2-, 4-, and 6-bit precisions from a shared NVFP4 base: the foundation from which other formats are composed. AutoFP6, the 6-bit E2M3 precision is obtained through mantissa refinement: by adding mantissa bits selectively to high quantization error base elements. The refinement metadata is stored separately from the base and applied on demand based on the runtime policy. Conversely, AutoFP2, the 2-bit precision is a quinary format, obtained by remapping the exponent bits of the base precision. 

\item \textit{System implementation} (\cref{sec:kernels}): We integrate AutoEMatic into vLLM with kernels that execute AutoFP using NVFP4 Tensor Cores and efficiently serve concurrent requests.

\item \textit{Hardware accelerator} (\cref{sec:hw}): We design an accelerator with an AutoFP-native mixed-precision datapath, which can process two AutoFP2 weights, one NVFP4 weight, or one AutoFP6 weight against a single streamed input activation.

\end{itemize}

To the best of our knowledge, AutoEMatic (\autoref{fig:teaser}) is the first framework to jointly adapt weight, activation, and KV-cache precision at runtime, across prefill and  decode, using an FP-based multi-scale format. Across reasoning and agentic benchmarks on models including Qwen3.8-27B, Gemma4-26B, and MiniMax-M2.7, AutoEMatic \textbf{improves accuracy by up to $\mathbf{21.7\%}$ over NVFP4} and \textbf{matches FP6$^{\text{g}}$} (the group-scaled 6-bit counterpart to NVFP4) on most benchmarks, using \textbf{$\mathbf{4.9}$ bits per weight} versus $4.5$ for NVFP4. AutoEMatic sustains this accuracy advantage while retaining \textbf{$\mathbf{90}$--$\mathbf{98\%}$ of NVFP4 throughput}. On custom hardware, AutoEMatic reduces latency by $1.6$--$2.3\times$ and energy by $7$--$23\%$ compared to an NVFP4-only datapath at iso-accuracy.
\section{Motivating Analyses}
\label{sec:motivation}

% We present experiments that motivate runtime precision adaptation and AutoEMatic's quantization format.

\subsection{Precision Demand Changes During Inference}
\label{sec:motivation_1}
To understand how the precision across layers of a model change throughout inference, we set up an oracle that determines for each layer at every prefill and decode step, the lowest precision (W$x$A$x$-KV$x$) such that the KL-divergence between the quantized and full-precision model's next-token distributions are within $10^{-3}$ nats, maintaining iso-generation. Following a full-precision generation, each layer is quantized sequentially, accounting for cumulative quantization effects. The oracle chooses between the following candidates: NVFP4, group-scaled 2-bit (E1M0) and 6-bit (E2M3) counterparts to NVFP4, and standard FP8 (E4M3). 

\noindent
\textbf{Decode Variations. }We apply this analysis to responses from DS-Qwen-14B on LiveCodeBench (LCB) and Qwen3.8-27B on $\tau^2$-bench, with up to 10K tokens per turn. \autoref{fig:motivation_1}(a) aggregates the oracle precisions assignments across layers within the early, middle, and late thirds of each response. The distribution varies across decoding stages and models, reflecting underlying layer-wise precision variations from one decode step to the next (\cref{app:layer_decode_precision}, Appx.).

\noindent
\textbf{Prefill Variations. }In agentic workloads, returns from tool calls extend the context and trigger additional prefills \cite{lu2026mix}, termed \textit{refills}. \autoref{fig:motivation_1}(b) shows Qwen3.8-27B's layer-wise oracle precision assignments over the initial prefill and refills. Demand changes sharply between prefill stages: refill 3 uses 2-bit for nearly all layers, while others demonstrate a broader precision distribution (\cref{sec:appendix_motivation}).

Therefore, precision demand depends jointly on the model, prompt, layer, and inference stage, and can change with any of these factors. The challenge of runtime-precision adaptation is to identify these dynamic variations without access to the full-precision reference used by the oracle.

\subsection{Limitation of Static Mixed-Precision Assignment}
\label{sec:motivation-static}
We further ask whether a static assignment calibrated in-distribution (calibrated and evaluated on the same dataset) can closely track the oracle's precision assignments. We examine two schemes: ModelOpt AutoQuantize \cite{nvidia2025modelopt}, which fixes per-layer precisions using calibrated sensitivities, and FlexQuant \cite{liu2025flexquant}, which follows a static schedule of lowering precision during inference. Both use 100 LiveCodeBench calibration samples under a 6-bit budget, matching the oracle's achieved average precision, and are evaluated on the remaining 780 samples. \autoref{fig:motivation_1}(c) reports underprovisioning: the fraction of decode steps where the oracle requires higher precision for a layer than the static policy assigns. AutoQuantize under-provisions $26.9\%$ on average and FlexQuant $37.6\%$. In-distribution calibration tailors to the aggregate precision demands of the calibration samples. However, the resulting fixed assignments or schedules do not respond to fine-grained variations across prompts and decode steps, leading to under-provisioning even on held-out samples from the same dataset, which directly impacts accuracy (\cref{sec:results}).

\begin{figure}[t]
    \centering
    \includegraphics[width=0.8\columnwidth, keepaspectratio]{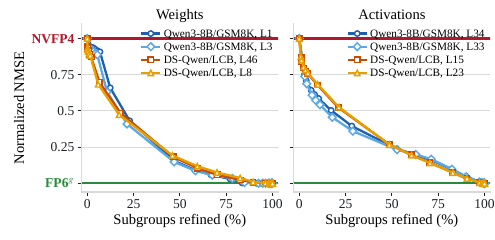}
    \vspace{-5mm}
    \caption{\textbf{Refining a subset of elements achieves NMSE comparable to 6-bit}. Y-axis is NMSE, such that NVFP4 is 1 and FP6$^{\text{g}}$ is 0, vs.\ fraction of subgroups refined to 6-bit.} % have only nmse instead of captured and have a horizontal line to show nvfp6 nmse
    \label{fig:motivation_2}
    \vspace{-6mm}
\end{figure}

\subsection{Non-uniform precision requirement within groups}
\label{sec:motivation_selective_refinement}
Raising the precision (refining) of a tensor, for example from 4 to 6 bits, is conventionally applied uniformly \cite{ramachandran2024clamp}. We instead ask whether the error of a group-scaled format such as NVFP4 is concentrated in a few elements of each group, so that refining those elements alone suffices. We partition each 16-element group into four 4-element subgroups ($SG = 4$; \cref{app:abl_subgroup_size} (Appx.)) and sweep NMSE thresholds from $0.3$ to $0.001$, and map the elements of every subgroup above the threshold to 6-bit (E2M3). For Qwen3-8B and DS-Qwen-14B (\autoref{fig:motivation_2}), refining $\sim50\%$ of weight subgroups achieves $\sim90\%$ of the NMSE reduction obtained by moving from NVFP4 to FP6$^{\text{g}}$ uniformly. Activations exhibit a similar trend. AutoFP leverages this to reduce metadata overhead when constructing higher precisions from NVFP4.

\section{AutoEMatic Quantization}
\label{sec:algorithm_methodology}

\subsection{Runtime Precision Adaptation Policy}
\label{sec:runtime_precision_adaptation}

\begin{figure}[t]
    \centering
    \includegraphics[width=\columnwidth, keepaspectratio]{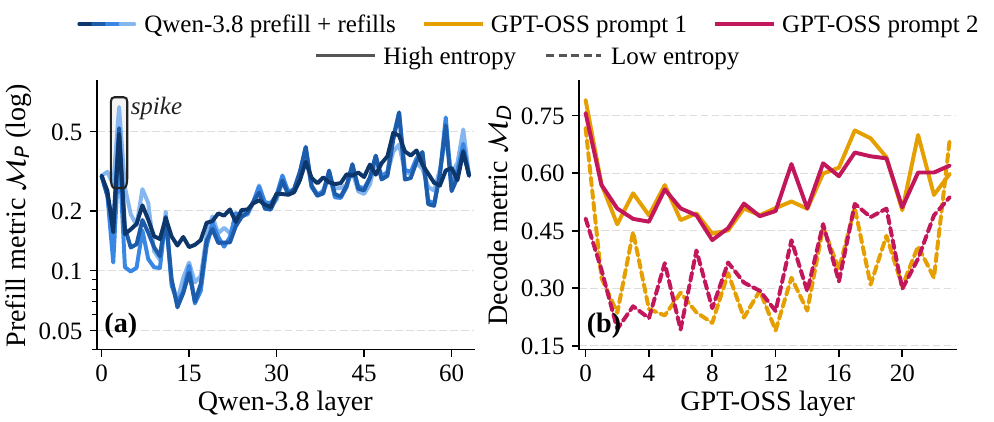}
        \vspace{-9mm}
        \caption{\textbf{Layer-wise behavior of $\mathcal{M}_P$ and $\mathcal{M}_D$.} {(a)} Prefill $\mathcal{M}_P$ of Qwen3.8-27B on a $\tau^2$-bench task. {(b)} Decode $\mathcal{M}_D$ of GPT-OSS-20B on RULER-16K, at high- and low-$\mathcal{G_E}$ steps.}
        \label{fig:signals}
        \vspace{-6mm}
\end{figure}

\noindent
\textbf{Precision-Adaptation Signals. }AutoEMatic determines precision at runtime from two complementary model-internal signals. The first is a global token-generation uncertainty signal $\mathcal{G_E}$, measured as the normalized Shannon entropy of the top-$K$ tokens' output logits (\cref{app:signals}). When the model is uncertain about the next token (high entropy), small quantization induced perturbations in the output logits can change which token it generates \cite{lee2026reset, lee2026polestar}. Such steps warrant higher precision and vice-versa. However, a global signal alone cannot determine layer-wise precision variations. The second signal is based on the layer structure of LLMs. LLM layers have a residual connection that carries its input directly to its output, with the attention and MLP blocks adding their contributions along the way \cite{zhao2026dynamicptq}. A layer whose output hidden states $\mathbf{H}_{\ell}(t)$ are close in representation to its input $\mathbf{H}_{\ell-1}(t)$ contributes little beyond copying its input to the output; we term it a low-activity layer, and its weights, activations, and KV cache can tolerate lower precision. This local layer-update activity $\mathcal{L_A}$ is measured at every step $t$, for every layer $l$ as,
\begingroup%
\setlength{\abovedisplayskip}{3pt}\setlength{\belowdisplayskip}{3pt}%
\begin{equation}
  \mathcal{L_A}^{\ell}(t) = \frac{\lVert \mathbf{H}_{\ell}(t) - \mathbf{H}_{\ell-1}(t) \rVert_2}{\lVert \mathbf{H}_{\ell-1}(t) \rVert_2 + \varepsilon},
  \label{eq:layer_activity}
\end{equation}
\endgroup%
where $\varepsilon$ ensures numerical stability. The two signals combine into one precision-demand metric for prefill or decode (P/D), $\mathcal{M}^{\ell}_{P/D}(t) = \lambda_{P/D}\, \mathcal{L_A}^{\ell}(t) + (1 - \lambda_{P/D})\, \mathcal{G_E}(t)$, with $\lambda_{P/D} \in [0, 1]$ set separately for each phase\footnote{$\mathcal{M}^{\ell}_{P/D}$ is a joint notation, $\mathcal{M}^{\ell}_{P}$ in prefill and $\mathcal{M}^{\ell}_{D}$ in decode.}. Rather than calibrated thresholds, AutoEMatic tracks the running mean $\mu$ and standard deviation $\sigma$ of $\mathcal{M}_{P/D}$ and establishes a set of rules, that vary across inference stages: sustained increases (decreases) of the metric over a fixed number of preceding layers relative to $\mu$ and $\sigma$ raise (lower) the precision of the next layer. Intuitively, these rules observe sustained trends in the metric as evidence that precision should change, rather than reacting to isolated fluctuations. Each layer executes at W$x$A$x$-KV$x$ with $x \in \mathcal{P} = \{2, 4, 6\}$ chosen by the rule. 2-bit activations and KV cache degrade accuracy substantially, therefore, the 2-bit level applies to weights only (W2A4-KV4). $\mathcal{P}$ denotes nominal precision levels; the policy is agnostic to the choice of levels or quantization format.

\noindent
\textbf{Prefill policy. }Prefill precedes token generation, so no token entropy exists and we set $\lambda_p = 1$ ($\mathcal{M}_P = \mathcal{L_A}$). We observe that, across models and datasets, $\mathcal{M}_P$ shows a {single large spike\footnote{Characterized by a sudden rise in the metric value.}} at an early layer, at a position largely invariant across prompts and refills (\autoref{fig:signals}(a)). Consistent with early-exit work~\cite{shen2026river}, the spike marks the layer by which the model has integrated the prompt context, so the layers up to and including it are the most sensitive and must execute at $\max \mathcal{P} = 6$ bits. Because the spike position differs across models, AutoEMatic detects it online as $\mathcal{M}_P^{\ell} > \mu + 3\sigma$. The subsequent layers follow the rule, $n_p$ layers below (above) $\mu$ lower (raise) the precision of the next layer by one level in $\mathcal{P}$, saturating at its bounds.

\begin{table}[t]
\centering
\vspace{-3mm}
\caption{\textbf{Runtime precision adaptation policy.}}
\label{tab:policy}
\renewcommand*{\arraystretch}{1.05}
\setlength\tabcolsep{4pt}
\resizebox{0.8\columnwidth}{!}{%
\begin{tabular}{l|c|c|c}
\Xhline{2\arrayrulewidth}
\rowcolor[HTML]{D3D3D3} \textbf{Policy} & \textbf{Prefill} & \multicolumn{2}{c}{\textbf{Decode}} \\
\rowcolor[HTML]{D3D3D3} & & High entropy & Low entropy \\
\Xhline{2\arrayrulewidth}
Metric  & $\mathcal{M}_P = \mathcal{L_A}$ & \multicolumn{2}{c}{$\mathcal{M}_D = 0.7\, \mathcal{L_A} + 0.3 \mathcal{G_E}$} \\
\hline
Initial precision & 6-bit & 4-bit & 4-bit \\
\hline
% Spike behavior & Layer after spike to 4-bit & \multicolumn{2}{c}{Replay spiking layer at $\max \mathcal{P}$} \\
\hline
Rules activate & After spike & \multicolumn{2}{c}{Layer 0} \\
\hline
Raise precision & $n_p$ layers $> \mu$ & $\beta$ layers $> \mu$ & $\alpha$ layers $> \mu$ \\
\hline
Lower precision & $n_p$ layers $< \mu$ & $\alpha$ layers $< \mu$ & $\beta$ layers $< \mu$ \\
\Xhline{2\arrayrulewidth}
\end{tabular}}
\vspace{-5mm}
\end{table}

\begin{figure}[t]
    \centering
    \includegraphics[width=0.8\linewidth, keepaspectratio]{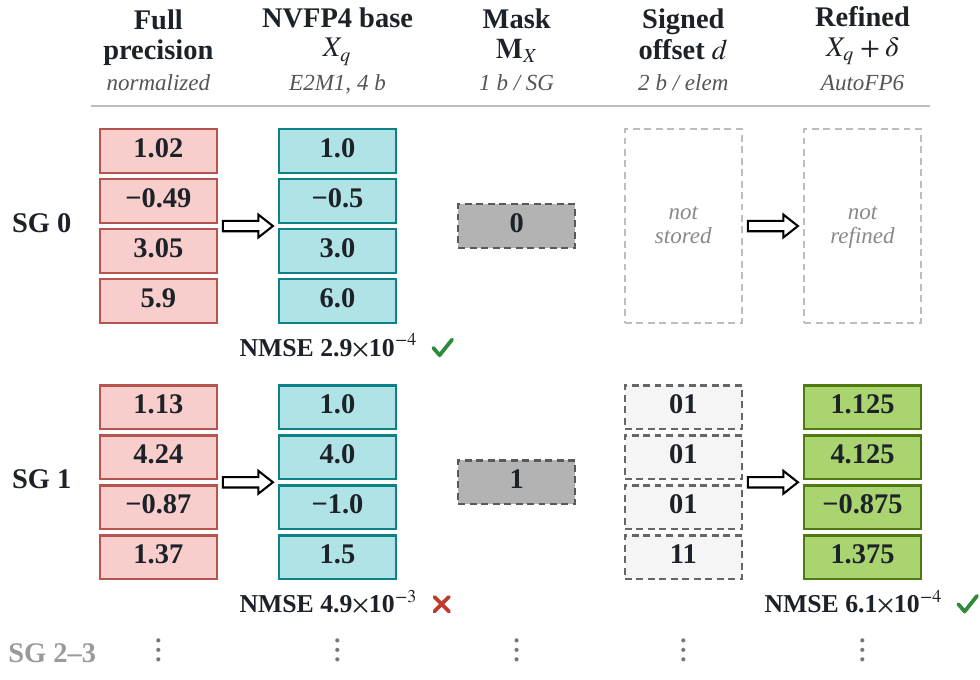}
        \vspace{-4mm}
        \caption{\textbf{Mantissa refinement of NVFP4 to AutoFP6.}}
        % ADd what we see here
        \label{fig:nvfp4_r6}
        \vspace{-7mm}
\end{figure} 

\noindent
\textbf{Decode policy. }During decode, we set $\lambda_D = 0.7$ incorporating generation entropy into the metric (\autoref{fig:signals}(b)). Since token entropy varies little between consecutive steps \cite{ramachandran2026thinkv} and the $\mathcal{G_E}$ of the current step is known only after its token is generated, $\mathcal{M}_D$ reuses $\mathcal{G_E}$ from the preceding step. Every decode step starts layer 0 at 4 bits and applies the rules in \autoref{tab:policy}. Since layer tolerance to quantization varies with generation entropy, we separate decoding into \emph{high-entropy} ($\mathcal{G_E} \ge \tau$) and \emph{low-entropy} ($\mathcal{G_E} < \tau$) steps by a threshold $\tau$. At high-entropy steps, the rule requires $\alpha$ consecutive low-activity layers to lower precision, but only $\beta$ consecutive high-activity layers to raise it, with $\alpha > \beta$. That is, AutoEMatic lowers precision more conservatively. At low-entropy, thresholds are reversed. 

% \vspace{-2mm}
\subsection{AutoFP Quantization Format}
\label{sec:quantization_format}

Storing one quantized copy per level in $\mathcal{P}$ would cost $13.5$ bits per weight. AutoFP instead stores one representation, the NVFP4 base with its group scales and additional metadata, and derives the 2- and 6-bit levels from it on demand.

\noindent
\textbf{AutoFP6. }Each additional mantissa bit adds local precision around every NVFP4 value. However, a floating-point value cannot be refined by `appending' mantissa bits. Since NVFP4 rounds to nearest, the full-precision value may lie below the E2M1 value, whereas appended mantissa bits can only select values above it, and not the nearest E2M3 value in the neighborhood that minimizes quantization error (e.g., $1.4$ is stored as $1.5$ in E2M1; appended bits can then reach only $1.5$–$1.875$, never $1.375$).

Therefore, instead of storing the additional metadata as mantissa bits, AutoFP encodes them as \emph{additive signed offsets}. Because the E2M1 grid is a subset of the E2M3 grid, every E2M1 value $X_q$ is also an E2M3 value. Refining $X_q$ only requires storing the direction and number of E2M3 grid steps to move from it. Since E2M3 adds three additional values around every $X_q$, a 2-bit two's-complement \underline{signed offset} $d \in \{-2, -1, 0, +1\}$ offers sufficient resolution. The step size around $X_q$ with exponent $e$ is $2^{e-3}$, and the refinement term is $\delta=d \cdot 2^{e-3}$ (the mantissa equivalent value of the signed offset) and the refined value is thus, $X_q + \delta$. The refinement is thus \underline{additive}. Since the step $2^{e-3}$ is dependent on the exponent of $X_q$, decoding $d$ exactly would require a per-element exponent computation (\cref{sec:ablations}). AutoFP therefore decodes $d$ with a fixed step $2^{-3}$, $\delta = d \cdot 2^{-3} \in \{-\tfrac{1}{4}, -\tfrac{1}{8}, 0, +\tfrac{1}{8}\}$. The fixed step is E2M3-like since it reproduces the E2M3 grid for values $<2$ and applies a step finer than E2M3 for larger values.

To reduce the metadata overhead of storing signed offsets, AutoFP6 refines only the subgroups with the largest NMSE reduction (\cref{sec:motivation_selective_refinement}). Each 16-element group carries a 4-bit refinement mask $\mathbf{M}_X$ with one bit per 4-element subgroup, set to 1 for a refined subgroup and 0 otherwise, and the signed offsets are stored only for the subgroups with mask bit of 1 (\autoref{fig:nvfp4_r6}).  The NVFP4 base is left intact, the signed offsets are stored separately and applied on demand for AutoFP6 layers. Since the refinement is additive, every refined operand then splits as $\widehat{\mathbf{X}} = \mathbf{X}_q + \mathbf{M}_X \odot \mathbf{X}_r$ for $\mathbf{X} \in \{\mathbf{W}, \mathbf{A}, \mathbf{K}, \mathbf{V}\}$, where $\mathbf{X}_r=\delta$. By linearity, a refined GEMM is, $\widehat{\mathbf{W}}\widehat{\mathbf{A}} = \mathbf{W}_q\mathbf{A}_q + (\mathbf{M}_W \odot \mathbf{W}_r)\mathbf{A}_q + \mathbf{W}_q(\mathbf{M}_A \odot \mathbf{A}_r) + (\mathbf{M}_W \odot \mathbf{W}_r)(\mathbf{M}_A \odot \mathbf{A}_r)$.

Two choices make this hardware-friendly. First, AutoEMatic imposes $N{:}4$ structured sparse \cite{ramachandran2025accelerating} refinement within each group, such that $N_W$ weight and $N_A=4-N_W$ activation subgroups are refined per group. Second, it keeps $\mathbf{M}_W$ and $\mathbf{M}_A$ disjoint. $\mathbf{M}_W$ selects $N_W$ subgroups for refinement (along $K$-dimension) and is shared across the rows of $\mathbf{W}$, and $\mathbf{M}_A$, shared across the columns of $\mathbf{A}$. This ensures disjointness holds for every row--column pair in the GEMM. Disjointness eliminates the second-order product of refinements, which contributes negligibly to accuracy (\cref{sec:ablations}), reducing the number of additional operations required for a refined layer. $\mathbf{M}_A$ is selected at runtime and $\mathbf{M}_W$ is selected offline under a refinement budget $\rho$, the fraction of weight subgroups refined, either by NMSE or by a refinement-aware GPTQ calibration that assigns refinement for subgroups with the largest Hessian-weighted error (\cref{app:gptq_calibration}).

\noindent
\textbf{AutoFP2. }Truncating the stored NVFP4 to 2 bits yields an E1M0 format representing ${\pm1, \pm2}$, which cannot represent zero and causes significant accuracy degradation. AutoFP2 instead is a quinary format, $Q \in \{-2, -1, 0, 1, 2\}$, derived from the NVFP4 code by retaining the sign bit $s$ and remapping the two exponent bits $e_1 e_0$ to a 2-bit magnitude in $\{0, 1, 2\}$ on demand via $\{e_1 \cdot e_0,\; e_1 \oplus e_0\}$. Nominally, representing AutoFP2 requires 3-bits. However, we show in \cref{sec:hw}, the accelerator packs two AutoFP2 digits into 5 bits, 2.5 bits per digit. AutoFP2 reuses the NVFP4 scales which normalize to $[-6, 6]$; since $Q$ reaches only $\pm2$, the scale is multiplied by a constant $c$ (\cref{sec:results}). As AutoFP2 is derived from NVFP4 on demand, it saves no memory bandwidth, and its main benefit is in prefill throughput and energy (\cref{sec:hw}).

\vspace{-2mm}
\section{System Implementation}
\vspace{-2mm}
\label{sec:kernels}
We implement kernels that executes AutoFP6 natively on NVFP4 Tensor Cores from a single packed model: E2M1 base values with E4M3 group scales $s_{W/A}$, signed offsets $d$ stored as one byte per refined subgroup, and refinement masks $\mathbf{M}_{W/A}$ stored as 4 bits per group along the $K$-dimension (\cref{app:memory_layout}, Appx.). Since Tensor Cores do not support sub-4-bit formats, AutoFP2 layers are emulated as NVFP4, which preserves the accuracy of the precision decision but not the performance of AutoFP2 execution.

\subsection{Refinement as an Extended-K NVFP4 Operation}
\label{sec:packed_refinement}

\begin{figure}[t]
    \centering
    \vspace{-1mm}
    \includegraphics[width=\columnwidth, keepaspectratio]{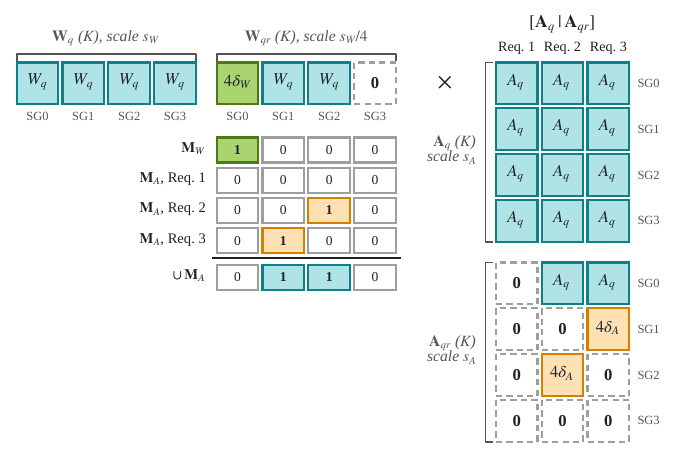}
    \vspace{-5.5mm}
    \caption{\textbf{Extended-$K$ GEMM for one 16-element group and three requests (batch size of 3).}}
    \label{fig:extended_k}
    \vspace{-8.5mm}
\end{figure}

A GEMM accumulates products along its reduction dimension $K$, so a sum of GEMMs with the same output shape equals one GEMM over operands concatenated along $K$ (see Appx.). AutoEMatic executes the refined GEMM of \cref{sec:quantization_format} in this manner. Since $\mathbf{M}_W$ and $\mathbf{M}_A$ are disjoint, the two refinement terms can be merged into a single GEMM of up to length-$K$ as, $\mathbf{W}_{qr}\mathbf{A}_{qr}$ (\autoref{fig:extended_k}). Each of these operands are constructed as follows: for subgroups that refine weights, $\mathbf{W}_{qr}$ holds the refinement term $\delta_W$ and $\mathbf{A}_{qr}$ holds the activation base values $A_q$ in the corresponding position, such that $\mathbf{W}_{qr}\mathbf{A}_{qr}$ realizes $(\mathbf{M}_W \odot \mathbf{W}_r)\mathbf{A}_q$. Similarly, for subgroups with activation-refinement, $\mathbf{W}_{qr}$ holds the weight base values $W_q$ and $\mathbf{A}_{qr}$ holds $\delta_A$, realizing $\mathbf{W}_q(\mathbf{M}_A \odot \mathbf{A}_r)$. The reduction dimension is up to length-$K$ because it depends on the number of subgroups refined. A refined layer thus executes as a single NVFP4 GEMM of reduction length up to $2K$ as, $[\mathbf{W}_q \,|\, \mathbf{W}_{qr}]\cdot[\mathbf{A}_q \,|\, \mathbf{A}_{qr}]$, which we term the \emph{extended-$K$ GEMM}. The GEMMs use Blackwell's \texttt{tcgen05.mma} instruction, which consumes only E2M1 values with E4M3 group scales. The refinement $\delta$ is not representable in E2M1 as is, whereas its scaled version $4\delta \in \{-1,-0.5, 0, +0.5\}$ is. The refined subgroups of $\mathbf{W}_{qr}$ and $\mathbf{A}_{qr}$ therefore hold $4\delta$, and the additional factor of four is absorbed into the scale of $\mathbf{W}_{qr}$, which is set to $s_W/4$. Each refined subgroup then contributes $(s_W/4 \cdot 4\delta_W)(s_A A_q) = s_W s_A\,\delta_W A_q$ or $(s_W/4 \cdot W_q)(s_A \cdot 4\delta_A) = s_W s_A\,W_q\delta_A$. The kernel fuses the construction of both extended operands with the MMA. Attention follows the same construction and runs on FlashInfer's FP8 attention kernel.
% A refined KV-cache token occupies twice the memory of an NVFP4 token, since its signed offsets are stored as a second NVFP4 value for the attention kernel. 
As in ThinKV~\cite{ramachandran2026thinkv}, AutoEMatic uses two page types of equal byte size: a base page holds $T$ unrefined tokens with only the base values, and a refinement page holds $T/2$ refined tokens with base values and signed offsets in NVFP4. This enables one paged allocator to serve both types.

% A refinement-aware page loader reads both page types and presents a uniform layout to the attention kernel.

\subsection{Serving Mixed-Precision Batches}
\label{sec:runtime_integration}

Requests in a batch may select different precisions for the same layer. The weight operands do not vary with that choice: $\mathbf{W}_q$ is shared by every request, and $\mathbf{W}_r$ with its mask $\mathbf{M}_W$ is fixed offline. Only the activation masks $\mathbf{M}_A$ and their signed offsets $d$ differ across requests. Rather than running a separate GEMM for each precision decision, AutoEMatic forms the \emph{union of masks} $\cup\mathbf{M}_A$, the per-group OR of the activation masks of all requests in the batch, and serves the whole batch with one extended-$K$ GEMM against a single $\mathbf{W}_{qr}$ built from it. As shown in \autoref{fig:extended_k}, $\mathbf{W}_{qr}$ holds $4\delta_W$ in its weight-refined subgroups, $W_q$ in every subgroup of $\cup\mathbf{M}_A$, and zero elsewhere. Each request then constructs its own $\mathbf{A}_{qr}$: $A_q$ in the weight-refined subgroups, $4\delta_A$ in the subgroups it refines.

% Requests in a batch may select different precisions for the same layer. The weight operands are nevertheless fixed, since $\mathbf{W}_q$,
% $\mathbf{W}_r$, and $\mathbf{M}_W$ are determined offline and a request's precision decides only whether $\mathbf{W}_r$ is applied; only the
% activation masks vary. AutoEMatic therefore builds one $\mathbf{W}_{qr}$ per layer from the \emph{union of masks} of all requests: $4\delta_W$ in the
% weight-refined subgroups, $W_q$ in every subgroup that any request refines in its activations, and zero elsewhere (\autoref{fig:extended_k}). Each
% request constructs its own $\mathbf{A}_{qr}$, which is zero for a 4-bit request, so requests of different precisions execute as one GEMM.

\providecommand{\bits}[1]{{\scriptsize(#1)}}
\begin{table}[t]
\centering
\caption{\textbf{Accuracy comparison of AutoEMatic and baselines across reasoning and agentic workloads.} Execution precision appears in parentheses next to accuracy. LB~v2: LongBench~v2, LCB: LiveCodeBench, LME: LongMemEval.}
\label{tab:main_results}
\renewcommand*{\arraystretch}{1.05}
\setlength\tabcolsep{2pt}
\resizebox{0.9\columnwidth}{!}{%
\begin{tabular}{ll|ccc|cc}
\Xhline{2\arrayrulewidth}
\rowcolor[HTML]{D3D3D3} & & \multicolumn{3}{c|}{\textbf{Reasoning}} & \multicolumn{2}{c}{\textbf{Agentic}} \\
\rowcolor[HTML]{D3D3D3} \multirow{-2}{*}{\textbf{Model}} & \multirow{-2}{*}{\textbf{Method }(budget)} & LB~v2 & AIME & LCB & $\tau^2$-bench & LME \\
\Xhline{2\arrayrulewidth}
\multirow{6}{*}{\rotatebox[origin=c]{90}{Qwen3-8B}}
 & \cellcolor[HTML]{EBEBEB}BF16 & \cellcolor[HTML]{EBEBEB}35.4 & \cellcolor[HTML]{EBEBEB}56.7 & \cellcolor[HTML]{EBEBEB}66.5 & \cellcolor[HTML]{EBEBEB}30.7 & \cellcolor[HTML]{EBEBEB}42.8 \\ \cline{2-7}
 & NVFP4 & 33.3 & 50.0 & 49.2 & 17.5 & 26.0 \\
 & FP6$^{\text{g}}$ & \textbf{35.4} & \textbf{56.7} & \textbf{61.4} & \textbf{30.7} & \textbf{39.9} \\
 & AutoQuantize \bits{4.5} & \textbf{35.4} & 50.0 & 60.0 & 20.2 & 31.9 \\
 & DP-LLM$^\dagger$ \bits{4.5} & 21.6 \bits{4.8} & 16.7 \bits{4.5} & 26.6 \bits{4.6} & 10.6 \bits{4.5} & 18.5 \bits{4.5} \\
 & \cellcolor[HTML]{D5E8D4}\textbf{AutoEMatic} & \cellcolor[HTML]{D5E8D4}\textbf{35.4} \bits{4.6} & \cellcolor[HTML]{D5E8D4}\textbf{56.7} \bits{4.7} & \cellcolor[HTML]{D5E8D4}60.5 \bits{4.8} & \cellcolor[HTML]{D5E8D4}27.2 \bits{4.6} & \cellcolor[HTML]{D5E8D4}38.2 \bits{5.1} \\
\Xhline{1\arrayrulewidth}
\multirow{6}{*}{\rotatebox[origin=c]{90}{Qwen3.8-27B}}
 & \cellcolor[HTML]{EBEBEB}BF16 & \cellcolor[HTML]{EBEBEB}56.9 & \cellcolor[HTML]{EBEBEB}93.3 & \cellcolor[HTML]{EBEBEB}76.3 & \cellcolor[HTML]{EBEBEB}74.6 & \cellcolor[HTML]{EBEBEB}82.8 \\ \cline{2-7}
 & NVFP4 & 48.7 & \textbf{86.7} & 70.6 & 50.0 & 70.6 \\
 & FP6$^{\text{g}}$ & \textbf{56.9} & \textbf{86.7} & \textbf{76.3} & \textbf{71.7} & \textbf{82.8} \\
 & AutoQuantize \bits{4.8} & \textbf{56.9} & \textbf{86.7} & 74.9 & 62.0 & 81.2 \\
 & DP-LLM$^\dagger$ \bits{4.8} & 25.4 \bits{4.7} & 56.7 \bits{4.8} & 51.4 \bits{4.5} & 25.6 \bits{4.6} & 15.9 \bits{4.7} \\
 & \cellcolor[HTML]{D5E8D4}\textbf{AutoEMatic} & \cellcolor[HTML]{D5E8D4}\textbf{56.9} \bits{4.5} & \cellcolor[HTML]{D5E8D4}\textbf{86.7} \bits{4.6} & \cellcolor[HTML]{D5E8D4}\textbf{76.3} \bits{4.6} & \cellcolor[HTML]{D5E8D4}\textbf{71.7} \bits{5.1} & \cellcolor[HTML]{D5E8D4}\textbf{82.8} \bits{4.8} \\
\Xhline{1\arrayrulewidth}
\multirow{6}{*}{\rotatebox[origin=c]{90}{Gemma-4-26B}}
 & \cellcolor[HTML]{EBEBEB}BF16 & \cellcolor[HTML]{EBEBEB}46.7 & \cellcolor[HTML]{EBEBEB}86.7 & \cellcolor[HTML]{EBEBEB}85.7 & \cellcolor[HTML]{EBEBEB}61.4 & \cellcolor[HTML]{EBEBEB}72.6 \\ \cline{2-7}
 & NVFP4 & 40.2 & 63.3 & 75.3 & 46.5 & 63.6 \\
 & FP6$^{\text{g}}$ & \textbf{45.5} & \textbf{80.0} & \textbf{83.1} & \textbf{60.5} & \textbf{70.1} \\
 & AutoQuantize \bits{5.0} & 41.0 & \textbf{80.0} & 80.1 & 58.6 & 68.1 \\
 & DP-LLM$^\dagger$ \bits{5.0} & 35.6 \bits{4.8} & 50.0 \bits{5.4} & 68.6 \bits{5.4} & 11.5 \bits{4.9} & 16.0 \bits{4.9} \\
 & \cellcolor[HTML]{D5E8D4}\textbf{AutoEMatic} & \cellcolor[HTML]{D5E8D4}\textbf{45.5} \bits{5.0} & \cellcolor[HTML]{D5E8D4}\textbf{80.0} \bits{5.0} & \cellcolor[HTML]{D5E8D4}\textbf{83.1} \bits{5.0} & \cellcolor[HTML]{D5E8D4}\textbf{60.5} \bits{5.1} & \cellcolor[HTML]{D5E8D4}\textbf{70.1} \bits{4.9} \\
\Xhline{2\arrayrulewidth}
\end{tabular}}
\vspace{-5mm}
\end{table}

\begin{figure}[t]
    \centering
    \includegraphics[width=\columnwidth, keepaspectratio]{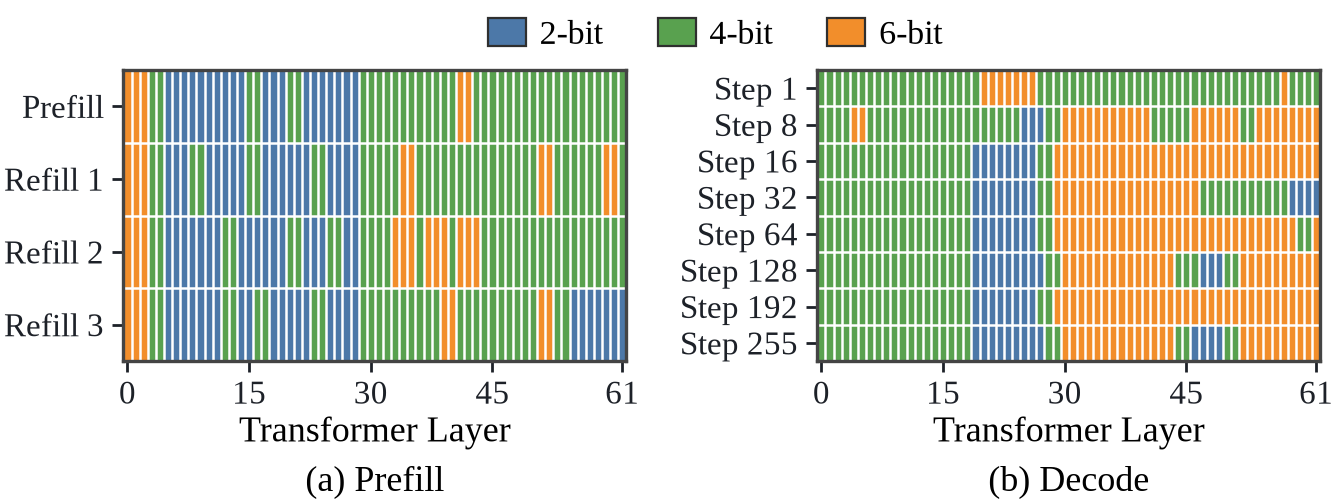}
    \vspace{-9mm}
    \caption{\textbf{Layer-wise precision assignment by AutoEMatic on MiniMax-M2.7.}}
    \label{fig:minimax_swe}
    \vspace{-8mm}
\end{figure}

\section{Accuracy and GPU Performance}
\label{sec:results}

\subsection{Experimental Setup}

\noindent
\textbf{Models, Datasets, and Baselines. }We evaluate Qwen3-8B, Qwen3.8-27B, DeepSeek-R1-Llama-70B, Gemma-4-26B-A4B, and MiniMax-M2.7 on reasoning (LongBench~v2~\cite{bai2025longbench}, AIME 2024~\cite{maa_aime_2024}, LiveCodeBench~\cite{jain2025livecodebench}) and agentic ($\tau^2$-bench~\cite{barres2025tau2bench}, LongMemEval~\cite{wu2024longmemeval}, SWE-bench Verified~\cite{openai2024swebenchverified}) workloads. For W$x$A$x$-KV$x$ quantization, we compare against uniformly quantized NVFP4 and FP6$^{\text{g}}$, AutoQuantize~\cite{nvidia2025modelopt}, which selects between NVFP4 or FP6$^{\text{g}}$ per layer under a bit budget, and DP-LLM~\cite{kwon2026dp}, which switches among 3- to 6-bit of an Any-Precision format~\cite{park2024any}. Since DP-LLM adapts only weights, we extend its policy to activations and KV cache (DP-LLM$^\dagger$) for a fair comparison. AutoQuantize, DP-LLM$^\dagger$, and AutoEMatic's refinement-aware GPTQ for weights use 256 samples from WikiText-2 for calibration. All experiments run in vLLM~\cite{kwon2023efficient} on NVIDIA GB300 GPUs with a 32K-token generation limit at temperature 0.5.

\noindent
\textbf{Hyperparameters. }We set $\mathcal{P} = \{2, 4, 6\}$; rules $n_p = 2$ and $\alpha = 4$, $\beta = 2$; entropy threshold $\tau = 0.10$, $\mathcal{G_E}$ is computed over the top-$K = 10$ logits; subgroup size $SG = 4$; AutoFP2 scale multiplier $c=2$; activation NMSE threshold $0.003$; weight refinement budget $\rho = 0.2$; KV page size $T = 64$. Hyperparameters are ablated in Appx.

\noindent
\textbf{Storage and Execution Precision. }\emph{Storage precision} is the packed model size in average bits per weight including metadata. For AutoQuantize, it is the target bit budget plus 0.5-bits for scale, 6-bits for DP-LLM$^\dagger$ and $4 + 0.5 + 2\rho = 4.9$ bits for AutoEMatic at $\rho = 0.2$ (since the masks are shared across rows for weights, it is negligible). \emph{Execution precision} is the nominal bitwidth at which layers execute averaged over layers and inference steps.

\subsection{Main Accuracy Results}

\noindent
\textbf{Comparison with Static and Runtime Baselines. }AutoEMatic operates without a bit budget: its execution precision follows the dynamic decisions of the runtime policy. For a fair comparison, AutoQuantize and DP-LLM$^\dagger$ are calibrated to the budget nearest AutoEMatic's achieved average. As shown in \autoref{tab:main_results}, across models and datasets, at $4.5$--$5.1$ bits of average execution precision, accuracy of AutoEMatic exceeds AutoQuantize by up to $10\%$ and DP-LLM$^\dagger$ by up to $66.9\%$ under a similar budget. AutoQuantize has a high underprovision rate which directly impacts its accuracy. AutoEMatic's advantage is particularly pronounced in agentic tasks, whose repeated prefills and long decodes impose dynamic precision demands, that DP-LLM$^\dagger$'s offline-calibrated runtime-policy cannot reliably track. DP-LLM$^\dagger$ is further penalized by its nested integer format: the sub-6-bit precisions extracted from the shared representation incur substantial error. Overall, through runtime adaptation driven by model-internal signals and the AutoFP format, \textbf{AutoEMatic improves accuracy by up to $21.7\%$ over NVFP4} and {matches FP6$^{\text{g}}$}, with a storage precision of 4.9 bits.
  
% MiniMax-M2.7 on SWE-bench Verified: execution precision with accuracy, as a wrapped table.
\begin{wraptable}{r}{0.5\columnwidth}
\vspace{-9mm}
\centering
\caption{\textbf{Accuracy comparison of MiniMax-M2.7 on SWE-bench Verified.}}
\label{tab:minimax_results}
\renewcommand*{\arraystretch}{1.05}
\setlength\tabcolsep{3pt}
\resizebox{\linewidth}{!}{%
\begin{tabular}{l|cc|c}
\Xhline{2\arrayrulewidth}
\rowcolor[HTML]{D3D3D3} & \multicolumn{2}{c|}{\textbf{Exec.\ Precision}} & \\
\rowcolor[HTML]{D3D3D3} \multirow{-2}{*}{\textbf{Method}} & Prefill & Decode & \multirow{-2}{*}{\textbf{Acc.\ (\%)}} \\
\Xhline{2\arrayrulewidth}
\cellcolor[HTML]{EBEBEB}FP8 & \cellcolor[HTML]{EBEBEB}8 & \cellcolor[HTML]{EBEBEB}8 & \cellcolor[HTML]{EBEBEB}71.0 \\
\hline
NVFP4 & 4 & 4 & 67.8 \\
FP6$^{\text{g}}$ & 6 & 6 & 70.5 \\
\cellcolor[HTML]{D5E8D4}\textbf{AutoEMatic} & \cellcolor[HTML]{D5E8D4}4.2 & \cellcolor[HTML]{D5E8D4}4.5 & \cellcolor[HTML]{D5E8D4}\textbf{70.5} \\
\Xhline{2\arrayrulewidth}
\end{tabular}}
\vspace{-5mm}
\end{wraptable}

\noindent
\textbf{Large-Scale Agentic Analysis. }On MiniMax-M2.7, a 230B-parameter agentic coding model, evaluated on SWE-bench Verified, AutoEMatic matches FP6$^{\text{g}}$ accuracy and remains within $0.5\%$ of baseline (at FP8) at an average execution precision of 4.2 bits in prefill and 4.5 in decode, against 6 bits for FP6$^{\text{g}}$ (\autoref{tab:minimax_results}). \autoref{fig:minimax_swe} demonstrates AutoEMatic's runtime adaptation across prefill and decode.

\vspace{-2mm}
\subsection{GPU Performance Analysis}
\label{sec:gpu_perf}
\begin{figure}[t]
    \centering
    \includegraphics[width=0.95\columnwidth, keepaspectratio]{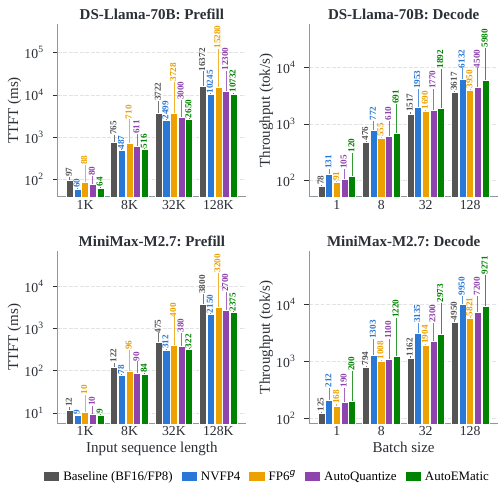}
    \vspace{-6mm}
    \caption{\textbf{Decode throughput and prefill TTFT analysis on $\mathbf{4\times}$ GB300 GPUs for DS-LLaMA-70B and MiniMax-M2.7.}}
    \label{fig:gpu_perf}
    \vspace{-5mm}
\end{figure}

\autoref{fig:gpu_perf} reports prefill time-to-first-token (TTFT) and decode throughput for DS-Llama-70B and MiniMax-M2.7 on $4\times$ GB300 GPUs, sweeping input lengths of 1K--128K tokens for prefill and batch sizes of 1--128 for decode with an 8K-token input and a 16K-token output. We compare against methods that execute on Tensor Cores: NVFP4 through the native CUTLASS W4A4 path, FP6$^\text{g}$ through a fused kernel that dequantizes E2M3 to FP8, and AutoQuantize uses either based on NVFP4/FP6$^\text{g}$ layers. AutoEMatic's runs use precision assignments from LongBench~v2.

\noindent
\textbf{Prefill. }AutoEMatic reduces TTFT by up to $1.4\times$ over FP6$^\text{g}$ and $1.25\times$ over AutoQuantize across input lengths. Specifically, despite AutoEMatic's runtime precision adaptation enabling superior accuracy compared to NVFP4, the TTFT is within $10\%$ of the NVFP4 baseline (small performance penalty). Two factors keep this performance penalty small: (a) when all requests in a batch choose 4-bit for a layer, they execute at NVFP4 throughput and (b) when any request chooses AutoFP6 it executes as a single extended-$K$ GEMM on Tensor Cores, instead of separate GEMMs (\cref{sec:ablations}). 

\noindent
\textbf{Decode. }AutoEMatic improves decode throughput by $1.5$--$2.5\times$ over the full-precision baseline across batch sizes, and attains $90$--$98\%$ of NVFP4 throughput, i.e., near-NVFP4 throughput at FP6$^\text{g}$ accuracy.

\section{Accelerator Architecture}
\vspace{-2mm}
\label{sec:hardware}
\label{sec:hw}

% The AutoEMatic accelerator demonstrates the upper bound of performance and energy efficiency achievable when the format is executed natively. 

\subsection{Architecture Overview}
\begin{figure}[t]
    \centering
    \includegraphics[width=0.70\columnwidth, keepaspectratio]{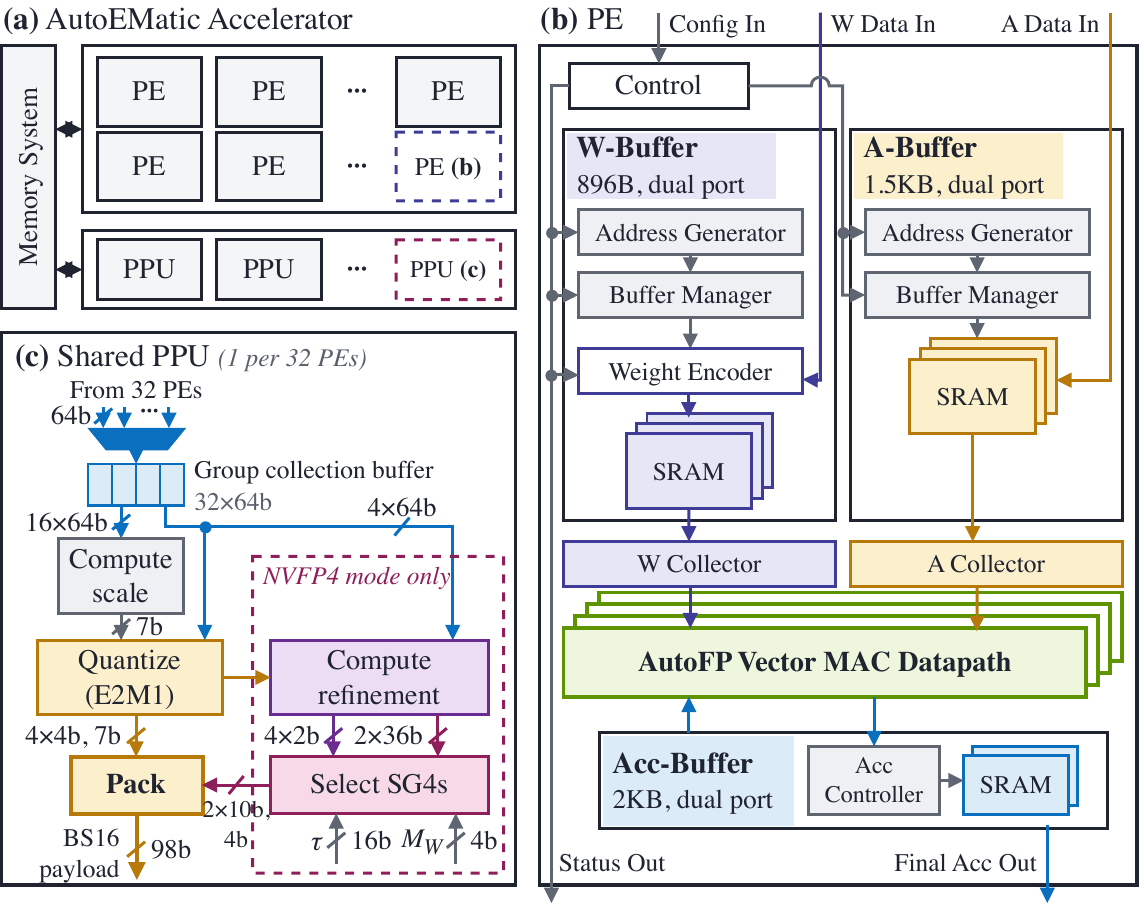}
    \vspace{-6mm}
    \caption{\textbf{AutoEMatic accelerator architecture.}}
    \vspace{-3mm}
    \label{fig:accelerator}
\end{figure}

\vspace{-2mm}
On GPUs, AutoFP2 is emulated and AutoFP6 requires an extension of the reduction dimension. An AutoFP-native datapath targets both limitations of existing hardware. We extend a representative accelerator MAGNet~\cite{venkatesan2019magnet,hooper2025fgmp} as proof-of-concept. The output-stationary local-weight-stationary (OS-LWS) design (\autoref{fig:accelerator}(a)) has 256 PEs and a global SRAM (L2) that loads data from DRAM (modeled). Each PE (\autoref{fig:accelerator}(b)) has mixed-precision vector MAC (vMAC) that computes one $V$-element NVFP4/AutoFP6 or two $V$-element AutoFP2 dot products per cycle. 
% Weight and activation buffers store NVFP4 bases and scales in separate banks from refinement masks and signed offsets for refined subgroups.
A PPU processes completed outputs (\autoref{fig:accelerator}(c)). Signals \texttt{mode\_nvfp4} and \texttt{refine\_en} select AutoFP2, NVFP4, or AutoFP6 modes of operation.
  
\begin{figure}[t]
    \centering
    \includegraphics[width=0.6\linewidth, keepaspectratio]{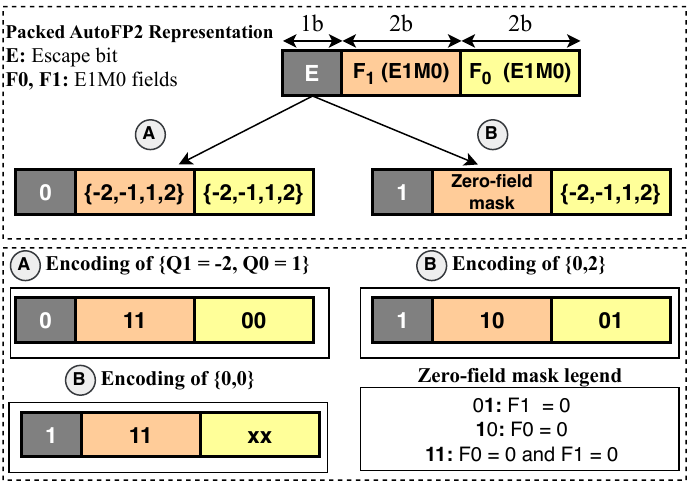}
        \vspace{-4mm}
        \caption{\textbf{AutoFP2 representation.}}
        \label{fig:quinary_format}
        \vspace{-7mm}
\end{figure}

\subsection{Weight Encoder and Packed AutoFP2 Format}
\label{sec:weight_encoder}
Weights are stored in DRAM as NVFP4 (\cref{sec:quantization_format}), but inside the PEs every weight is held in AutoFP2, so that the vMAC can execute all modes with AutoFP2 arithmetic units. Two requirements follow. First, for the vMAC to process AutoFP2 at twice the throughput of NVFP4, each weight load from the buffers must deliver two AutoFP2 weights. Second, in NVFP4 mode a single AutoFP2 digit cannot represent an NVFP4 value. We introduce a packed AutoFP2 format that satisfies both by encoding a pair of AutoFP2 digits $(Q_1, Q_0)$ in 5 bits (2.5 bits per digit). In the packed format (\autoref{fig:quinary_format}), an escape bit $E$ precedes two 2-bit fields $F_1$ and $F_0$. A nonzero AutoFP2 digit is one of $\pm1$ and $\pm2$. When both digits are nonzero ($E = 0$), $F_1$ and $F_0$ hold them in E1M0. Otherwise ($E = 1$), $F_1$ records which digits are zero and $F_0$ holds the nonzero digit, if any. The weight encoder forms the AutoFP2 packed format as weights enter the PE and writes them to the weight buffer. The packing overhead is amortized by the weight stationary dataflow. In AutoFP2 mode (\texttt{mode\_nvfp4=0}), it applies the exponent remapping of \cref{sec:quantization_format} to two weights from adjacent rows (different $K$ dimension) that share an input activation and packs the two digits. In NVFP4/AutoFP6 modes (\texttt{mode\_nvfp4=1}), the pair represents one NVFP4 weight as $W_q = c_0 Q_0 + c_1 Q_1$ with fixed coefficients. We set $W_q = (5Q_0 + 2Q_1)/2$, such that their linear combinations spans the maximum NVFP4 magnitude $6$ (derivation in Appx.). For e.g., $0.5 = (5\cdot1 + 2\cdot(-2))/2$. Each AutoFP2 digit is a Boolean function of the E2M1 bits (\cref{app:hw-encoding}).

\subsection{vMAC Datapath Design}
\begin{figure}[t]
      \centering
      \includegraphics[width=0.80\columnwidth, keepaspectratio]{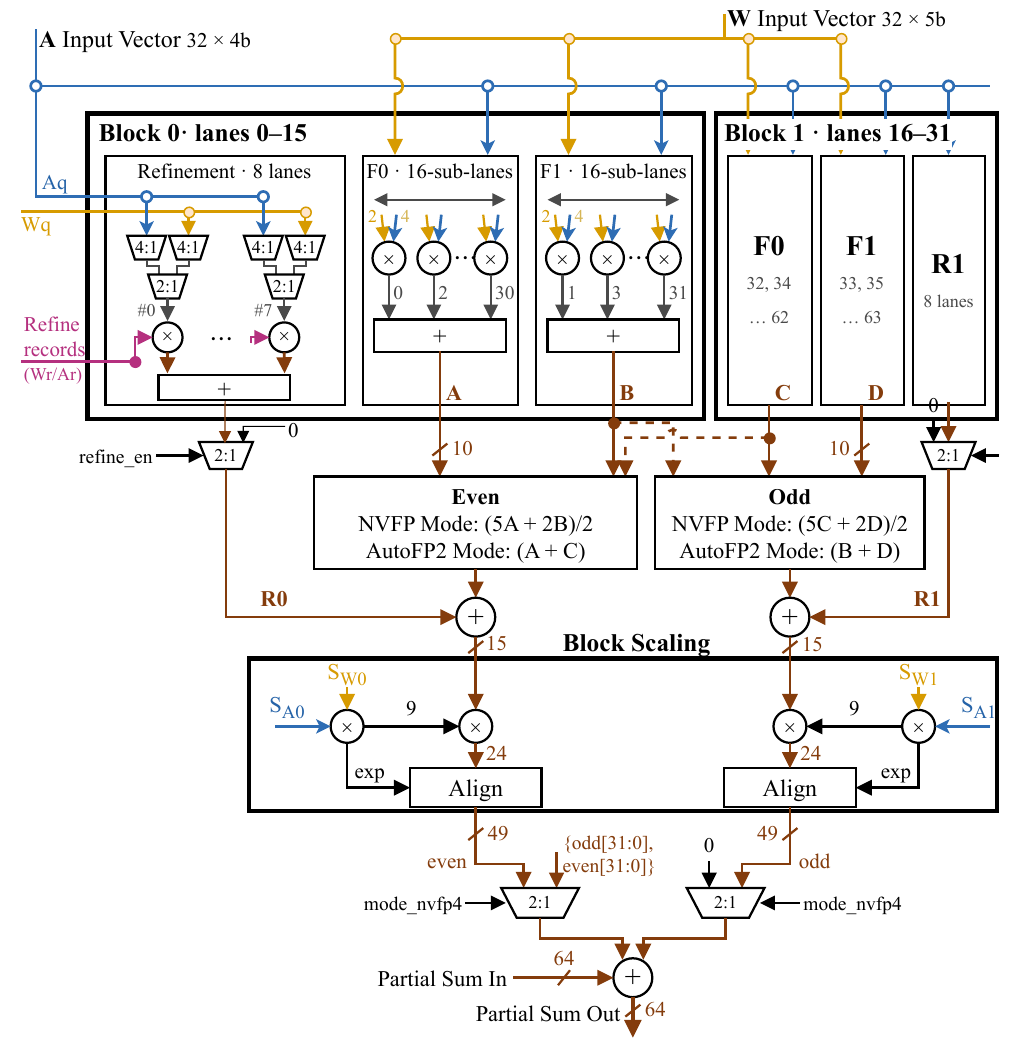}
      \vspace{-6mm}
      \caption{\textbf{AutoFP vector MAC datapath.}}
      \vspace{-7mm}
      \label{fig:pe_datapath}
  \end{figure}

\vspace{-2mm}
\label{sec:vmac}
Each PE is a $V=32$ element wide vMAC (\autoref{fig:pe_datapath}) organized as two $V/2=16$ base lane blocks each with dedicated set of refinement lanes and pipelined into multiply and accumulate stages. The vMAC accepts weights in the packed AutoFP2 format. The refinement data has a separate input interface carrying the masks and signed offsets of the refined subgroups. Each base lane $k$ comprises an even ($F_{0,k}$) and an odd ($F_{1,k}$) AutoFP2 sub-lane that shares one streamed NVFP4 activation. For energy-efficiency, all computations use fixed-point arithmetic (\cref{app:vmac_fixed_point}). 
% Each input activation enters as $U_k = 2A_q(k)$, since doubling an E2M1 value yields an integer~\cite{hayashi2026dgemm}. Similarly, in NVFP4 mode the packed format represents twice the NVFP4 weight, $2W_q = 5Q_0 + 2Q_1$.

\noindent
\textbf{Base Lanes. }In the multiply stage, the escape bit $E$ of each packed pair decodes the two digits. When $E = 1$, $F_1$ identifies which digit is zero, and the corresponding sub-lane is bypassed: its product is set to zero. For a nonzero digit, $F_1$ or $F_0$ holds an E1M0 code, $\pm1$ or $\pm2$, so each sub-lane forms $A_k F_{j,k}$, $j \in \{0, 1\}$, with simple shifters. Within a block, the even and odd sub-lanes feed separate adder trees in the adder stage, $A=\sum_{k=0}^{15}A_kF_{0,k}$ and $B=\sum_{k=0}^{15}A_kF_{1,k}$ for lanes 0--15, and $C$, $D$ for lanes 16--31. Every execution mode differs only in how the tree outputs are combined. In NVFP4 mode, the $A, B$ correspond to the partial sum accumulated from the same weight $W_{q,k}$, therefore the mode-dependent reduction unit computes $P0 = (5A + 2B)/2$ (implemented in fixed-point) to form the corresponding NVFP4 partial sum, and similarly for $P_1$, as shown in \autoref{fig:pe_datapath}. The base lanes perform the same operation in AutoFP6 mode. In AutoFP2 mode, within a block the two sub-lanes process corresponding weights from different rows (so that they can share one input activation). The weights of the same row are therefore spread across the even and odd sub-lanes of both blocks, so the mode-dependent reduction unit sums across blocks, $P_0 = A + C$ and $P_1 = B + D$.

% In the multiply stage, each sub-lane forms $U_k Q_{j,k}, j \in \{0,1\}$ through simple shifters. Within each block, the even and odd sub-lanes feed separate adder trees, $A=\sum_{k=0}^{15}U_kQ_{0,k}$ and $B=\sum_{k=0}^{15}U_kQ_{1,k}$ for lanes 0--15, and $C$, $D$ for lanes 16--31. The modes differ only in how the tree outputs are combined. In NVFP4 mode, the $A, B$ correspond to the partial sum accumulated from the same weight $2W_{q,k}$, so by distributivity, $P_0 = 5A + 2B = \sum_k U_k(5Q_{0,k} + 2Q_{1,k}) = \sum_k (2A_q)(2W_q)$ corresponds to four times the dot product of block 0 in fixed-point, and similarly for $P_1$, as shown in \autoref{fig:pe_datapath}. In AutoFP2 mode, even and odd sub-lanes of a block hold two different rows (so that they can share one input activation), hence partial sums are summed across blocks as $(2(A+C),;2(B+D))$, yielding two independent 32-element dot products.

% In AutoFP2 mode, even and odd sub-lanes of a block correspond to two different rows (this is necessary to enable the sub-lanes to share one input activation), therefore AutoFP2 partial sums are summed across blocks as, $(2(A+C),\;2(B+D))$, yielding two independent 32-element dot products. The factor of 2 for partial sums in AutoFP2 mode ensures AutoFP2 partial sums are also four times the dot product in fixed-point.

\noindent
\textbf{Sparse Refinement Lanes. }The refinement datapath adds $V/4 = 8$ refinement lanes per block as two $SG=4$-lane slots (in \cref{sec:ablations} we establish this as sufficient). In AutoFP2/NVFP4 mode (\texttt{refine\_en=0}), the refinement lanes are gated for energy efficiency. Because $\mathbf{M}_W$ and $\mathbf{M}_A$ are disjoint, a refined weight/activation subgroup occupies exactly one slot. The collector supplies each slot with the refinement mask, four 2-bit signed-offsets $d$, and a bit identifying weight/activation refinement. Each lane has two 4:1 muxes that pick the corresponding base-weight and activation subgroups using the mask; a 2:1 selector chooses the corresponding $X_q$: $A_q$ for weight refinement, or $W_q$ for activation refinement (note the similarity with construction of $W_{qr}, A_{qr}$). Each lane forms $R = 2X_q d$ in fixed-point arithmetic. Each refinement lanes' contributions are accumulated as $\mathcal{R} = \sum R$. It is accumulated with the base lanes partial sums as, $O_j = 4P_j + \mathcal{R}$. 

% Finally, $P_0$ and $P_1$ of base lanes are shifted left by two bits, to align with the refinement products, giving $F_j = 4P_j + \sum R$ at sixteen times the dot product. (for e.g. $A_q = 1.5$, $W_q = 3$, $d_W = -1$: $P = 18=(2\cdot1.5)(2\cdot3)$, $R = -3$, $F = 69 = 16 \cdot 1.5\,(3 - 1/8)$).

% Since $\delta = d \cdot 2^{-3}$, the FP equivalent is $X_q\delta = R/16$, therefore $R$ is sixteen times the product in fixed-point. Each refinement lanes' contributions are accumulated as $\sum R$. Finally, $P_0$ and $P_1$ of base lanes are shifted left by two bits, to align with the refinement products, giving $F_j = 4P_j + \sum R$ at sixteen times the dot product. (for e.g. $A_q = 1.5$, $W_q = 3$, $d_W = -1$: $P = 18=(2\cdot1.5)(2\cdot3)$, $R = -3$, $F = 69 = 16 \cdot 1.5\,(3 - 1/8)$).
  
\noindent
\textbf{Scaling and accumulation. }Each $O_j$ is multiplied by the scale factor of its corresponding block. In AutoFP2 mode, each $O_j$ is a 32-element dot product corresponding to two different weight rows with the same activations. In NVFP4 and AutoFP6 mode, each $O_j$ corresponds to a 16-element dot product along the same row (same $K$ dimension). After scaling, $O_0$ and $O_1$ are summed into one 64-bit accumulator in NVFP4/AutoFP6, while in AutoFP2 mode $O_0, O_1$ are held in the two 32-bit halves of the accumulator.

\begin{figure}[t]
    \centering
    \includegraphics[width=0.75\columnwidth, keepaspectratio]{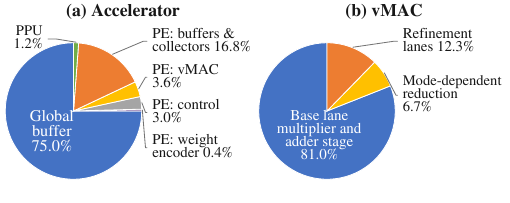}
    \vspace{-7mm}
    \caption{\textbf{Area breakdown of (a) the AutoEMatic accelerator and (b) its vMAC.}}
    \label{fig:area}
    \vspace{-7mm}
\end{figure}

\subsection{Shared Post-Processing Unit}
The PPU (\autoref{fig:accelerator}(c)) quantizes completed PE outputs into the next layer's input activations. If a layer requires refinement, it also computes per-subgroup NMSE and selects subgroups with NMSE above $\tau$ excluding those in $\mathbf{M}_W$ and assigns signed offsets $d$. The PPU also performs non-linear operations, and computes the metric $\mathcal{M}_{P/D}$.

\vspace{-2mm}
\subsection{Accelerator Evaluation}
\label{sec:accel_eval}
\vspace{-2mm}
\noindent
\textbf{Implementation. }The accelerator is implemented in SystemC, with RTL generated using Siemens Catapult HLS. We use a 2MB global SRAM, 896B weight, 1.5KB activation, and 2KB accumulation buffers. We perform synthesis in TSMC 3nm at a frequency of 1 GHz. Synopsys VCS is used for simulation and PrimeTime PX for power estimation. 

\noindent
\textbf{Area Analysis. }\autoref{fig:area}(a) breaks down the area of the AutoEMatic accelerator. The 2MB global SRAM occupies $75\%$ of the area. The PE array, with all 256 PEs, takes $23.8\%$. The per-PE weight encoders account for only $0.4\%$ and the vMAC $3.6\%$ of the total area; within a vMAC the refinement lanes ($12.3\%$) and the mode-dependent reduction logic for NVFP4- and AutoFP2-mode ($6.7\%$), result in a net $1.2\times$ overhead over an NVFP4-only vMAC. The accelerator is only $1.1\%$ more area than the NVFP4 baseline.

\begin{figure}[t]
    \centering
    \includegraphics[width=0.85\columnwidth, keepaspectratio]{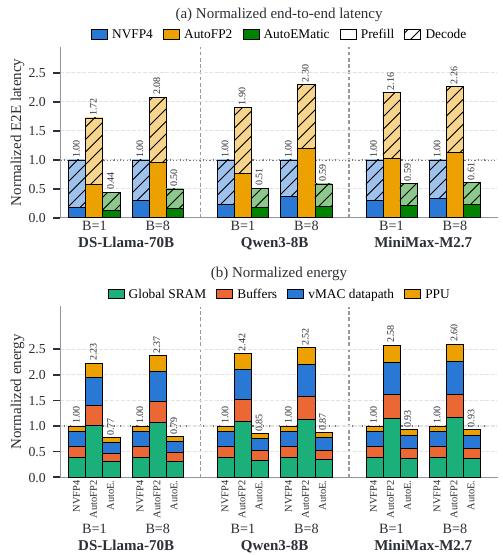}
    \vspace{-5mm}
    \caption{\textbf{Normalized latency and energy comparison with single-format baselines, under iso-accuracy.}}
    \label{fig:accel_perf}
    \vspace{-6mm}
\end{figure}

\begin{table}[t]
\centering
\vspace{-2mm}
\caption{\textbf{Component ablations} with DS-LLaMA-70B on LCB.}
\label{tab:ablation_components}
\renewcommand*{\arraystretch}{1.05}
\setlength\tabcolsep{3pt}
\resizebox{\columnwidth}{!}{%
\begin{tabular}{l|c|c|c}
\Xhline{2\arrayrulewidth}
\rowcolor[HTML]{D3D3D3} \textbf{Component} & \textbf{Bits} & \textbf{Acc.} & \textbf{2 / 4 / 6 (\%)} \\
\Xhline{2\arrayrulewidth}
\cellcolor[HTML]{EBEBEB}1.\ BF16 & \cellcolor[HTML]{EBEBEB}16 & \cellcolor[HTML]{EBEBEB}68.4 & \cellcolor[HTML]{EBEBEB}-- \\
\hline
2.\ NVFP4 (W-A-KV) & 4.5 & 51.7 & 0 / 100 / 0 \\
3.\ \quad + AutoEMatic runtime policy w/  $\mathcal{P}=\{2,4,6\}$ copies & 13.5 & 66.2 & 20 / 40 / 40 \\
\hline
4.\ AnyPrecision LLM on $\mathcal{P}$ + AutoEMatic runtime policy & 6 & 43.8 & 8 / 42 / 50 \\
\hline
\cellcolor[HTML]{D5E8D4}5.\ AutoFP6  & \cellcolor[HTML]{D5E8D4}4.9 & \cellcolor[HTML]{D5E8D4}64.8 & \cellcolor[HTML]{D5E8D4}0 / 0 / 100 \\
\cellcolor[HTML]{D5E8D4}6.\ \quad + Refinement-aware GPTQ & \cellcolor[HTML]{D5E8D4}4.9 & \cellcolor[HTML]{D5E8D4}{65.6} & \cellcolor[HTML]{D5E8D4}0 / 0 / 100 \\
\hline
\cellcolor[HTML]{D5E8D4}7.\ \textbf{AutoEMatic format + runtime policy}, $\mathcal{P}=\{2,4,6\}$ & \cellcolor[HTML]{D5E8D4}4.9 & \cellcolor[HTML]{D5E8D4}\textbf{65.6} & \cellcolor[HTML]{D5E8D4}15 / 45 / 40 \\
\Xhline{2\arrayrulewidth}
\end{tabular}}
\vspace{-8mm}
\end{table}

\noindent
\textbf{Iso-Accuracy Performance Comparison. }We compare against single-format AutoFP2 and NVFP4 baselines under iso-accuracy with AutoEMatic by replaying its layer-wise precision traces, as recorded on LiveCodeBench. Layers above a baseline's native precision execute with an extended reduction dimension, accumulating into the same partial sums; the NVFP4 design, for instance, executes a refined layer analogous to the extended-$K$ GEMM (\cref{sec:kernels}). \autoref{fig:accel_perf}(a) reports end-to-end latency normalized to the NVFP4 design at batch sizes 1 and 8. AutoEMatic is $1.6$--$2.3\times$ faster than NVFP4 for two reasons: AutoFP2 layers run at twice the NVFP4 throughput, which benefits compute-bound prefill most ($1.3$--$1.8\times$ faster), and refined layers execute natively in the sparse refinement lanes without the additional pass an extended-$K$ reduction requires. Across evaluated models, on average, AutoEMatic achieves a compute density of 5.46 TOPS/mm$^2$, compared to 3.23 TOPS/mm$^2$ for NVFP4 and 6.16 TOPS/mm$^2$ for AutoFP2.

\noindent
\textbf{Iso-Accuracy Energy Comparison. }\autoref{fig:accel_perf}(b) reports energy for the same traces, normalized to NVFP4 and broken down into different components, each including static and dynamic energy. AutoEMatic consumes $0.77$--$0.93\times$ the energy of NVFP4, as (a) all modes use AutoFP2 lanes and (b) lower runtime. Its vMAC datapath averages 15fJ/op, against 22fJ/op for NVFP4 and 10fJ/op for AutoFP2.

\vspace{-2mm}
\section[Discussion and Ablations]{Discussion and Ablations\protect\footnote{See Appx. for additional discussions and ablations.}}
\label{sec:ablations}
\begin{figure*}[t]
    \centering
    \includegraphics[width=\textwidth]{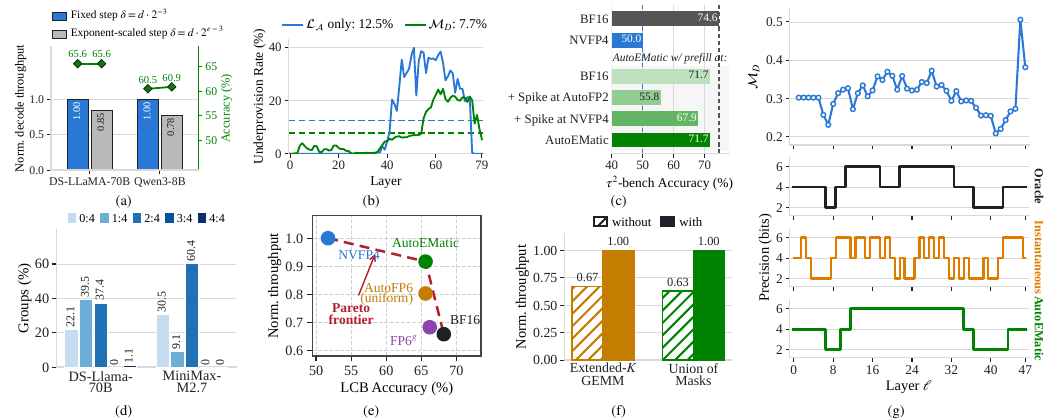}
    \vspace{-5mm}
    \caption{\textbf{Ablations.} \textbf{(a)} Fixed vs.\ exponent-scaled decoding of $d$. \textbf{(b)} Underprovision rate of AutoEMatic. \textbf{(c)} Prefill precision at spike. \textbf{(d)} Refined subgroups per group. \textbf{(e)} Accuracy vs.\ throughput pareto. \textbf{(f)} Kernel ablations. \textbf{(g)} For one decode step: $\mathcal{M}_D$ across layers and the precision assigned by the oracle, an instantaneous rule, and AutoEMatic.}
    \label{fig:ablations}
    \vspace{-5mm}
\end{figure*}
  
\noindent
\textbf{[\abltag{policy}{Algorithm}+\abltag{quant}{Quantization}] Component Ablations. }\autoref{tab:ablation_components} isolates the runtime policy and the quantization format. Applying AutoEMatic's runtime policy to separately quantized AutoFP2, NVFP4, and FP6$^{\text{g}}$ copies (row 3), attains $66.2\%$, but needs 13.5 stored bits per weight. Applied to the integer multi-scale Any-Precision format (row 4), attains only $43.8\%$. This suggests that a favorable accuracy--efficiency tradeoff requires both: the runtime policy and an FP-based multi-scale format. Combined (row 7), AutoEMatic preserves $65.6\%$, within $0.6\%$ of the separate-copy upper bound, at 4.9 bits, executing $40\%$ layers at 6-bit, $45\%$ at 4-bit and $15\%$ at 2-bit. Selecting the refined weight subgroups by NMSE alone, with a threshold of $0.003$ at the same budget $\rho$ (\autoref{tab:ablation_components}, row 5), attains $64.8\%$; refinement-aware GPTQ (row 6), which selects them by their reduction in the Hessian-weighted error raises accuracy to $65.6\%$.

\noindent
\textbf{[\abltag{quant}{Quantization}] Refinement Decoding. }\autoref{fig:ablations}(a) compares decoding signed offsets with the fixed step $2^{-3}$ against exact, per-element exponent computation $2^{e-3}$ on DS-LLaMA-70B and Qwen3-8B on LCB. The fixed-step has similar accuracy as the exact exponent computation, while achieving up to $28\%$ higher decode throughput. This is because the fixed-step does not require loading the base value when constructing the extended-$K$ GEMM operands.

\noindent
\textbf{[\abltag{policy}{Algorithm}] AutoEMatic Runtime Policy vs. Oracle. }\autoref{fig:ablations}(b) compares per-layer underprovision rate of AutoEMatic's precision decisions against the oracle of \cref{sec:motivation} restricted to the precision set $\mathcal{P}$. AutoEMatic underprovisions only $7.7\%$ of layer executions on average, and its 2/4/6-bit distribution of $15/45/40\%$ closely tracks the oracle's $20/41/39\%$. Using $\mathcal{L_A}$ alone has a higher rate of underprovisioning, demonstrating the combined metric $\mathcal{M}_D$ as the optimal choice. We omit $\mathcal{G_E}$ from this comparison because it is a global signal shared across layers. 

\noindent
\textbf{[\abltag{policy}{Algorithm}] Rule-based vs. Instantaneous Adaptation. }\autoref{fig:ablations}(g) traces one decode step of DS-Qwen-14B layer by layer: $\mathcal{M}_D$ (top) and the precision assigned by the oracle of \cref{sec:motivation_1}, by an instantaneous policy that switches with every rise and fall in $\mathcal{M}_D$, and by AutoEMatic's rule-based policy. Evident from the oracle's own behavior, the rules are not ad hoc, oracle precision does not rise (drop) immediately after an increase (decrease) in the metric, but only once the change persists over several layers. The instantaneous policy switches on blips and drops to 2 bits at layers where the oracle requires 6. AutoEMatic's rules are built inspired by this behavior of the oracle and change precision only once the model has provided sufficient evidence.
  
\noindent
\textbf{[\abltag{policy}{Algorithm}] Prefill Precision at the Spike. }\autoref{fig:ablations}(c) varies only how the spiking layer in prefill is executed, keeping the decode policy fixed. Running prefill entirely in BF16 gives $71.7\%$, the same as AutoEMatic, which executes the spike layers at 6 bits. Lowering the spike layer to NVFP4 costs $3.8\%$ and to AutoFP2 $15.9\%$. 

\noindent
\textbf{[\abltag{quant}{Quantization}] Subgroups Refined Per Group. }\autoref{fig:ablations}(d) shows the distribution of weight+activation subgroups refined per group, $N_W+N_A$, for DS-LLaMA-70B on LCB and MiniMax-M2.7 on SWE-bench. $22$--$31\%$ of groups need no refinement (0:4), and only $1.1\%$ of DS-LLaMA-70B groups and none in MiniMax-M2.7 exceed 2:4, supporting the choice of 8 refinement lanes.

\begin{figure}[t]
    \centering
    \includegraphics[width=\columnwidth]{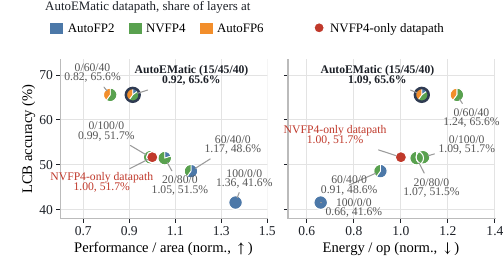}
    \vspace{-5mm}
    \caption{\textbf{DS-LLaMA-70B LCB accuracy vs.\ performance per area and energy per op}, normalized to an NVFP4-only datapath.}
    \vspace{-5mm}
    \label{fig:accel_mix}
\end{figure}

\noindent
\textbf{[\abltag{sys}{System}] Kernel Ablations. }\autoref{fig:ablations}(e) plots accuracy against decode throughput normalized to NVFP4 for DS-LLaMA-70B on LCB at batch 512. AutoEMatic lies on the Pareto frontier: $1.3\times$ faster than FP6$^{\text{g}}$, and $1.1\times$ faster than uniform AutoFP6. AutoEMatic's superior performance-accuracy tradeoff stems from the runtime policy: it assigns 4 bits to most layers, which then run at NVFP4 throughput, and 6 bits only where accuracy requires it, which still runs as a single extended-K GEMM on the same NVFP4 Tensor Cores. \autoref{fig:ablations}(f) ablates AutoEMatic's two kernel mechanisms with different batch sizes. Without the extended-$K$ GEMM, executing the base and refinement GEMMs separately and summing their outputs lowers throughput by $1.5\times$ at batch size 1; without the union of masks, splitting mixed-precision batches into per-precision sub-batches lowers throughput by $1.6\times$ at batch size 512.

\noindent
\textbf{[\abltag{hw}{Accelerator}] Precision Mix Sweep. }\autoref{fig:accel_mix} evaluates different layer-wise precision distributions on DS-LLaMA-70B against an NVFP4-only datapath. AutoEMatic's runtime assignment ($15/45/40$) lies on the Pareto frontiers of accuracy versus area efficiency and accuracy versus energy efficiency among the evaluated configurations of different precision distributions, achieving $65.6\%$ accuracy with only an $8\%$ performance/area penalty and $9\%$ higher energy/op.

% This penalty reflects the overhead associated with supporting AutoFP mixed-precision datapath.

% but its mixed-precision native datapath, 

% while pay

% Compared to the NVFP4 datapath, with AutoEMatic's own runtime assignment ($15/45/40$), raises accuracy to $65.6\%$, while achieving $92\%$ of the NVFP4-only performancearea at $1.09\times$ energy/op

%  Executing all layers at AutoFP2, AutoEmatic reaches $1.36\times$ the performance per area of the NVFP4-only design at $34\%$ lower energy per operation. With AutoEMatic's own runtime assignment ($15/45/40$), raises accuracy to $65.6\%$, while retaining $92\%$ of the NVFP4-only performance/area at $1.09\times$ energy/op.

\vspace{-2mm}
\section{Related Work}
\vspace{-2mm}
\label{sec:related_work}

\noindent
\textbf{Multi-scale quantization. }Any-Precision LLM~\cite{park2024any}, Matryoshka Quantization~\cite{nair2025matryoshka}, AnyBCQ~\cite{park2026anybcq}, and RRQ~\cite{luo2026rrq} store several precisions within one integer representation, as nested bit planes or successive residual stages, from which the selected precision is reassembled before computation.

\noindent
\textbf{Runtime precision adaptation. }FlexQuant~\cite{liu2025flexquant}, DP-LLM~\cite{kwon2026dp}, MoBiQuant~\cite{wang2026mobiquant}, and QuickSilver~\cite{khanna2025quicksilver} adapt weight precision during decoding through predetermined schedules, calibrated error thresholds, or trained routers, and Mix-Quant~\cite{lu2026mix} assigns NVFP4 to prefill and BF16 to decoding. For the KV cache, PM-KVQ~\cite{liu2026pmkvq} progressively lowers precision and ThinKV~\cite{ramachandran2026thinkv} assigns precision by reasoning importance. Appx. provides an extended discussion.

\vspace{-2mm}
\section{Conclusion}
\vspace{-2mm}
AutoEMatic shows that adapting precision to the evolving numerical demands of LLM inference improves its accuracy--efficiency tradeoff. AutoFP switches precision at runtime through mantissa refinement and exponent remapping of a shared NVFP4 base, without requiring redundant weight copies. Future work will explore achieving memory benefits with AutoFP2 and extending AutoFP to higher precisions through joint exponent and mantissa refinement.

\bibliography{main}
\bibliographystyle{mlsys2025}

%%%%%%%%%%%%%%%%%%%%%%%%%%%%%%%%%%%%%%%%%%%%%%%%%%%%%%%%%%%%%%%%%%%%%%%%%%%%%%%
%%%%%%%%%%%%%%%%%%%%%%%%%%%%%%%%%%%%%%%%%%%%%%%%%%%%%%%%%%%%%%%%%%%%%%%%%%%%%%%
% SUPPLEMENTAL CONTENT AS APPENDIX AFTER REFERENCES
%%%%%%%%%%%%%%%%%%%%%%%%%%%%%%%%%%%%%%%%%%%%%%%%%%%%%%%%%%%%%%%%%%%%%%%%%%%%%%%
%%%%%%%%%%%%%%%%%%%%%%%%%%%%%%%%%%%%%%%%%%%%%%%%%%%%%%%%%%%%%%%%%%%%%%%%%%%%%%%
\clearpage
\appendix
\startcontents[appendix]

\section*{\textbf{\large Appendix}}
\printcontents[appendix]{}{1}{%
  \setcounter{tocdepth}{2}
}

\section{Extended Related Work}
\label{app:related}

\noindent
\textbf{Static post-training quantization. }Most PTQ methods choose a precision offline and keep it fixed. GPTQ~\cite{frantar2023gptq} quantizes a weight matrix one column at a time and uses approximate inverse-Hessian information to update the remaining weights, compensating the quantization error. AWQ~\cite{lin2024awq} identifies salient weight channels from activation statistics and protects them by per-channel scaling before quantization.

\noindent
\textbf{Group-scaled floating-point formats and their extensions. }MX formats~\cite{rouhani2023microscaling,ocp2023mx} share a power-of-two E8M0 scale across a group of 32 FP8, FP6, FP4, or INT8 elements, and NVFP4~\cite{alvarez2025nvfp4} stores E2M1 elements in groups of 16 with an E4M3 group scale and a per-tensor FP32 scale. MX+~\cite{lee2025mxplus} observes that the group's largest element does not need its exponent field given the shared scale and repurposes those bits as extra mantissa for that element. M$^2$XFP~\cite{hu2026m2xfp} adds two metadata bits per subgroup, which serve as extra mantissa bits for the subgroup's largest activation and as a scale multiplier for weights. AdaMX~\cite{luo2026adamx} chooses per group between FP4 and INT4 grids combined with group-maximum mantissa extension or scale refinement at a constant bit width. These extensions are fixed after quantization and refine one element or one scale per group. AutoEMatic is generally applicable to any group-scaled floating-point format. 

\noindent
\textbf{Multi-scale quantization. }Multi-scale methods store several precisions in one representation so that precision can change without separate checkpoints. Any-Precision LLM~\cite{park2024any} quantizes a seed model at the lowest bit width with non-uniform clustering, then adds one bit at a time by splitting each cluster, storing the codes as bit planes so that lower-bit codes are prefixes of higher-bit codes. Matryoshka Quantization~\cite{nair2025matryoshka} co-trains a single INT8 model so that slicing its most significant bits yields usable INT4 and INT2 models. AnyBCQ~\cite{park2026anybcq} builds 2- to 4-bit models on binary-coded quantization by freezing existing bit planes and adding residual planes with re-optimized scales, and RRQ~\cite{luo2026rrq} represents weights as a 2-bit base plus successive 2-bit quantized residuals, so that 4-, 6-, and 8-bit precisions are sums of residual stages.

\noindent
\textbf{Runtime precision adaptation. }Mix-Quant~\cite{lu2026mix} applies NVFP4 to prefill and keeps BF16 for decoding, fixing each phase's precision. FlexQuant~\cite{liu2025flexquant} stores both INT8 and INT4 weights and, during decoding, replaces INT8 layers with INT4 in an order fixed offline by KL-divergence analysis whenever a moving average of output entropy falls below a threshold. MoBiQuant~\cite{wang2026mobiquant} decomposes weights into recursive 2-bit residual slices and uses a trained token router to decide how many slices each token activates. QuickSilver~\cite{khanna2025quicksilver} computes, at a layer, the entropy of each token's hidden representation and assigns 8-, 4-, or 2-bit precision by thresholds.

\noindent
\textbf{KV-cache compression. }KIVI~\cite{liu2024kivi} quantizes the key cache per channel and the value cache per token to 2 bits without tuning, and KVQuant~\cite{hooper2024kvquant} combines per-channel pre-RoPE key quantization, sensitivity-weighted non-uniform datatypes, and a dense-and-sparse split that keeps outliers separately. For long chain-of-thought reasoning, PM-KVQ~\cite{liu2026pmkvq} lowers the bit width of the KV cache progressively as the context grows and allocates higher bit widths to more sensitive blocks, and ThinKV~\cite{ramachandran2026thinkv} classifies chain-of-thought segments into thought types from attention sparsity, assigning token precision by thought importance and evicting tokens of less critical thoughts.

\noindent
\textbf{Mixed-precision accelerators. }MicroScopiQ~\cite{ramachandran2025microscopiq} retains outlier weights at higher precision and offsets their additional bits by pruning the least important weights, executing on multi-precision integer processing elements with a network-on-chip that routes the high-precision outliers. AdaMX~\cite{luo2026adamx} shares one multiplier datapath across its group formats and applies up to three per-group scalar correction terms through an auxiliary path. MixFP4~\cite{zou2026mixfp4} selects one of two FP4 element formats, E2M1 or E1M2, per group, signaled by the sign bit of the E4M3 group scale, and decodes both to a common internal E2M2 representation.

\begin{table}[t]
  \centering
  \caption{Low-precision floating-point formats.}
  \label{tab:low-precision-formats}
  \resizebox{\columnwidth}{!}{%
  \begin{tabular}{lccc}
    \toprule
    \rowcolor{tabgray} Format & S/E/M & Magnitudes & Bits/elem. \\
    \midrule
    E2M1 (NVFP4) & 1/2/1 & $\{0, 0.5, 1, 1.5, 2, 3, 4, 6\}$ & 4.5 \\
    E2M3 (FP6$^{\text{g}}$) & 1/2/3 & $0$--$7.5$ in steps of $\tfrac{1}{8}, \tfrac{1}{4}, \tfrac{1}{2}$ & 6.5 \\
    E1M0 & 1/1/0 & $\{1, 2\}$ & 2.5 \\
    E0M1 & 1/0/1 & $\{0, 1\}$ & 2.5 \\
    Quinary digit (AutoFP2) & -- & $\{0, 1, 2\}$ & 3.0 \\
    \midrule
    E4M3 (scale) & 1/4/3 & $2^{-9}$ -- $448$ & -- \\
    \bottomrule
  \end{tabular}}
  \vspace{-5mm}
\end{table}

\begin{table}[t]
  \centering
  \caption{Positive magnitudes of E2M1 (NVFP4) and E2M3 (FP6$^{\text{g}}$).}
  \label{tab:e2m1-e2m3}
  \renewcommand*{\arraystretch}{1.05}
  \setlength\tabcolsep{3pt}
  \resizebox{\columnwidth}{!}{%
  \begin{tabular}{c|cc|>{\columncolor[HTML]{D5E8D4}}cccc>{\columncolor[HTML]{D5E8D4}}cccc}
    \Xhline{2\arrayrulewidth}
    \rowcolor[HTML]{D3D3D3} & \multicolumn{2}{c|}{\textbf{E2M1 (NVFP4)}} & \multicolumn{8}{c}{\textbf{E2M3 (FP6$^{\text{g}}$)}} \\
    \rowcolor[HTML]{D3D3D3} \multirow{-2}{*}{$\mathbf{E \backslash M}$} & 0 & 1 & 000 & 001 & 010 & 011 & 100 & 101 & 110 & 111 \\
    \Xhline{2\arrayrulewidth}
    00 & 0 & 0.5 & 0 & 0.125 & 0.25 & 0.375 & 0.5 & 0.625 & 0.75 & 0.875 \\
    01 & 1 & 1.5 & 1 & 1.125 & 1.25 & 1.375 & 1.5 & 1.625 & 1.75 & 1.875 \\
    10 & 2 & 3   & 2 & 2.25  & 2.5  & 2.75  & 3   & 3.25  & 3.5  & 3.75 \\
    11 & 4 & 6   & 4 & 4.5   & 5    & 5.5   & 6   & 6.5   & 7    & 7.5 \\
    \Xhline{2\arrayrulewidth}
  \end{tabular}}
  \vspace{-5mm}
\end{table}

\section{Low-Precision Floating-Point Formats}
\label{appendix:number-formats}

We write E$e$M$m$ for a signed floating-point format with one sign bit, $e$ exponent bits, and $m$ mantissa bits. \Cref{tab:low-precision-formats} summarizes them.

\subsection{NVFP4}
\label{app:nvfp4}

NVFP4 \cite{alvarez2025nvfp4} stores each element as an E2M1 value (exponent bias 1) with magnitudes $\{0, 0.5, 1, 1.5, 2, 3, 4, 6\}$ under two levels of scaling. A tensor $X$ is split into groups $X_j$ of 16 contiguous elements along the reduction dimension; each group has an E4M3 \emph{group scale} $B_j$, and the tensor has an FP32 \emph{tensor scale} $T$:
\begin{align}
  T &= \frac{\operatorname{amax}(X)}{6 \cdot 448},
  &
  B_j &= Q_{\mathrm{E4M3}}\!\left(\frac{\operatorname{amax}(X_j)}{6\,T}\right),
  \label{eq:nvfp4-scales}\\
  q_i &= Q_{\mathrm{E2M1}}\!\left(\frac{x_i}{T B_j}\right),
  &
  \widehat{x}_i &= T B_j\, q_i, \quad i \in X_j,
  \label{eq:nvfp4-quant}
\end{align}
where $Q_F$ rounds to the nearest value of format $F$ (ties to even). The group scale maps each group's absolute maximum onto $6$, the largest E2M1 value, and the tensor scale maps the largest group scale onto $448$, the largest E4M3 value. With one 8-bit scale per 16 elements, NVFP4 costs $4 + 8/16 = 4.5$ bits per element.

\subsection{FP6$^{\text{g}}$}
\label{app:nvfp6}

FP6$^{\text{g}}$, the group-scaled 6-bit counterpart of NVFP4 used as a baseline throughout the paper, keeps the group size and scaling hierarchy of NVFP4 and replaces E2M1 by E2M3 (exponent bias 1). The two additional mantissa bits give 32 magnitudes, in steps of $\tfrac{1}{8}$ on $[0, 2)$, $\tfrac{1}{4}$ on $[2, 4)$, and $\tfrac{1}{2}$ on $[4, 7.5]$, and lower the worst-case relative rounding error of normal values from $1/4$ to $1/16$. The scales follow \cref{eq:nvfp4-scales,eq:nvfp4-quant} with the E2M3 maximum $7.5$ in place of $6$, so FP6$^{\text{g}}$ costs $6 + 8/16 = 6.5$ bits per element. \Cref{tab:e2m1-e2m3} lists the positive magnitudes of both element formats.

\subsection{Two-Bit Formats and the AutoFP2 Digit}
\label{app:two-bit}

With one bit for the sign, a 2-bit floating-point format spends its other bit on either the exponent or the mantissa. \textbf{E1M0} spends it on the exponent: giving the magnitudes $\{1, 2\}$ and the codebook $\{\pm 1, \pm 2\}$, which cannot represent zero, so small values are forced to $\pm 1$. \textbf{E0M1} spends it on the mantissa: both codes are subnormal, giving the ternary codebook $\{-1, 0, +1\}$ with a redundant $-0$, which represents zero but only one nonzero magnitude. AutoFP2's quinary digit $\{-2, -1, 0, 1, 2\}$ keeps both by remapping the two exponent bits of the NVFP4 base (\cref{sec:quantization_format}).

\subsection{Sharing NVFP4 Scales with AutoFP2}
\label{app:scale-sharing}

A tensor stored in NVFP4 that executes at AutoFP2 reuses the NVFP4 scales $T$ and $B_j$. These normalize each group to $[-6, 6]$, whereas AutoFP2 digits reach only $\pm 2$, so decoding with $T B_j$ alone would shrink each group by up to $3\times$. The scale is therefore multiplied by a scale multiple $c$:
\begin{equation}
  \widehat{x}_i = c\, T B_j\, q_i, \qquad q_i \in \{-2, -1, 0, 1, 2\}.
  \label{eq:two-bit-dequant}
\end{equation}
Setting $c = 3$ matches the two ranges exactly, with reconstruction levels $\{0, \pm 3, \pm 6\}$ in NVFP4 units. We instead use $c = 2$, with levels $\{0, \pm 2, \pm 4\}$, for two reasons. First, a factor of two is hardware-friendly. Second, the AutoFP2 digit $1$ absorbs the four E2M1 magnitudes $\{1, 1.5, 2, 3\}$, which hold most group-normalized values. Reconstructing this digit at $2$ rather than $3$ places it in the middle of that range and halves its worst-case error from $2$ to $1$ NVFP4 units.

\subsection{Worked Example of Mantissa Refinement}
\label{app:refinement-example}

\Cref{tab:refinement-example} applies mantissa refinement to one 16-element NVFP4 group with scale $s = T B_j = 0.25$.

\begin{table}[t]
  \centering
  \caption{Mantissa refinement of one NVFP4 group ($s = 0.25$). Values are in NVFP4 units; codes are sign\,|\,exponent\,|\,mantissa. Green rows are the refined subgroups (mask \texttt{0110}).}
  \label{tab:refinement-example}
  \renewcommand*{\arraystretch}{1.05}
  \setlength\tabcolsep{3pt}
  \resizebox{\columnwidth}{!}{%
  \begin{tabular}{cr|ccc|cccc}
    \Xhline{2\arrayrulewidth}
    \rowcolor[HTML]{D3D3D3} & & \multicolumn{3}{c|}{\textbf{NVFP4 base}} & \multicolumn{4}{c}{\textbf{AutoFP6 refinement}} \\
    \rowcolor[HTML]{D3D3D3} \multirow{-2}{*}{\textbf{Subgroup}} & \multirow{-2}{*}{$x/s$} & $q$ & Code & Err. & $\delta$ & Code & $q{+}\delta$ & Err. \\
    \Xhline{2\arrayrulewidth}
     & 1.00 & 1.0 & 0\,01\,0 & 0 & -- & -- & -- & -- \\
     & $-$0.50 & $-$0.5 & 1\,00\,1 & 0 & -- & -- & -- & -- \\
     & 3.00 & 3.0 & 0\,10\,1 & 0 & -- & -- & -- & -- \\
    \multirow{-4}{*}{SG0} & 6.00 & 6.0 & 0\,11\,1 & 0 & -- & -- & -- & -- \\
    \hline
    \rowcolor[HTML]{D5E8D4}  & 1.13 & 1.0 & 0\,01\,0 & 0.13 & $+\tfrac{1}{8}$ & \texttt{01} & \textbf{1.125} & \textbf{0.005} \\
    \rowcolor[HTML]{D5E8D4}  & 2.24 & 2.0 & 0\,10\,0 & 0.24 & $+\tfrac{1}{8}$ & \texttt{01} & \textbf{2.125} & \textbf{0.115} \\
    \rowcolor[HTML]{D5E8D4}  & $-$0.87 & $-$1.0 & 1\,01\,0 & 0.13 & $+\tfrac{1}{8}$ & \texttt{01} & \textbf{$-$0.875} & \textbf{0.005} \\
    \rowcolor[HTML]{D5E8D4} \multirow{-4}{*}{SG1} & 1.37 & 1.5 & 0\,01\,1 & 0.13 & $-\tfrac{1}{8}$ & \texttt{11} & \textbf{1.375} & \textbf{0.005} \\
    \hline
    \rowcolor[HTML]{D5E8D4}  & 2.76 & 3.0 & 0\,10\,1 & 0.24 & $-\tfrac{1}{4}$ & \texttt{10} & \textbf{2.75} & \textbf{0.010} \\
    \rowcolor[HTML]{D5E8D4}  & $-$0.62 & $-$0.5 & 1\,00\,1 & 0.12 & $-\tfrac{1}{8}$ & \texttt{11} & \textbf{$-$0.625} & \textbf{0.005} \\
    \rowcolor[HTML]{D5E8D4}  & 4.24 & 4.0 & 0\,11\,0 & 0.24 & $+\tfrac{1}{8}$ & \texttt{01} & \textbf{4.125} & \textbf{0.115} \\
    \rowcolor[HTML]{D5E8D4} \multirow{-4}{*}{SG2} & $-$2.02 & $-$2.0 & 1\,10\,0 & 0.02 & $0$ & \texttt{00} & \textbf{$-$2.0} & \textbf{0.020} \\
    \hline
     & 0.52 & 0.5 & 0\,00\,1 & 0.02 & -- & -- & -- & -- \\
     & $-$1.98 & $-$2.0 & 1\,10\,0 & 0.02 & -- & -- & -- & -- \\
     & 0.00 & 0.0 & 0\,00\,0 & 0 & -- & -- & -- & -- \\
    \multirow{-4}{*}{SG3} & 2.95 & 3.0 & 0\,10\,1 & 0.05 & -- & -- & -- & -- \\
    \Xhline{2\arrayrulewidth}
  \end{tabular}}
\vspace{-5mm}
\end{table}

\begin{figure}[t]
    \centering
    \includegraphics[width=\columnwidth, keepaspectratio]{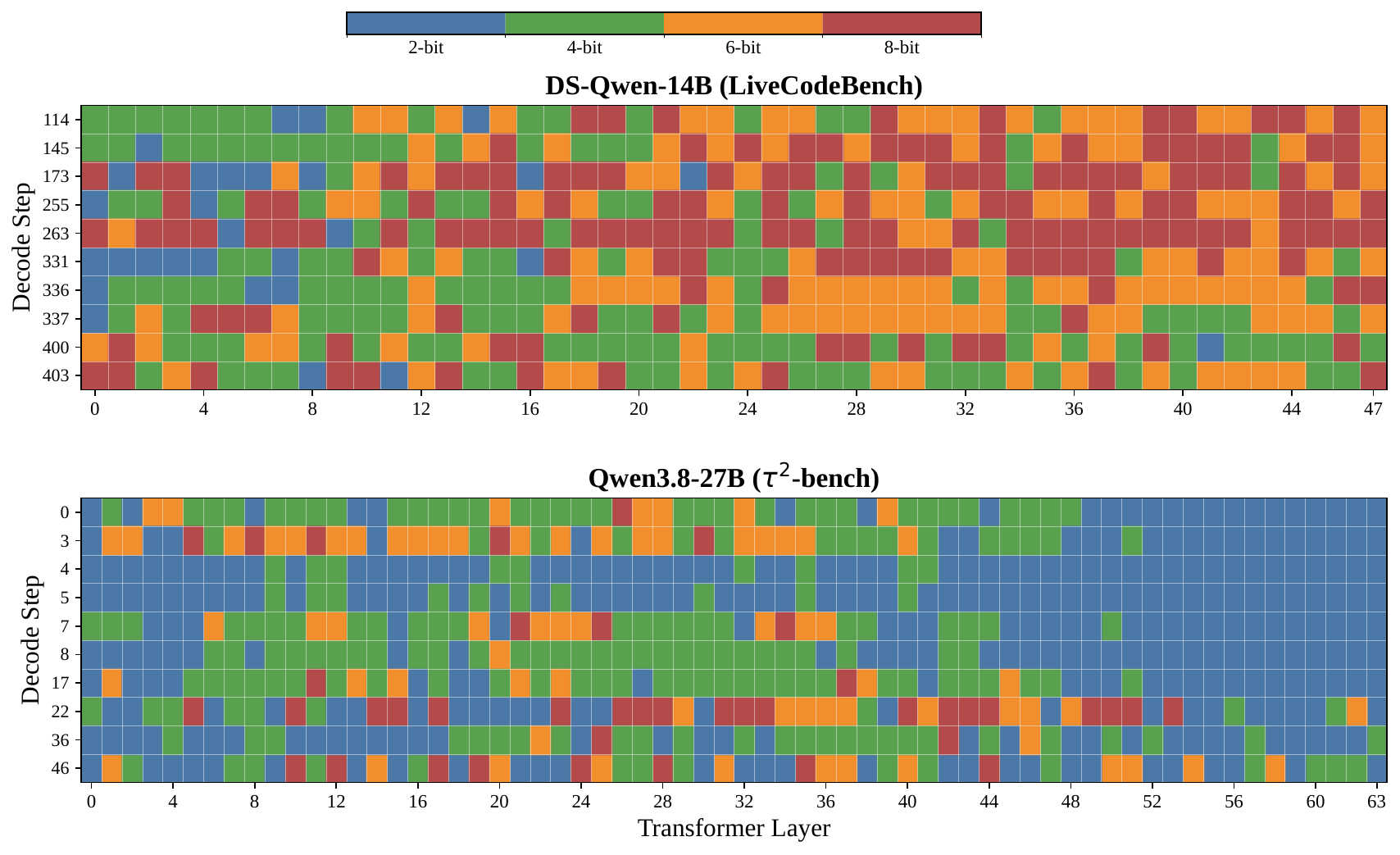}
    \vspace{-6mm}
    \caption{\textbf{Layer-wise precision across decode steps.} Per-layer precision at ten decode steps of one DS-Qwen-14B LiveCodeBench response (top) and one Qwen3.8-27B $\tau^2$-bench response (bottom).}
    \label{fig:layer_decode_heatmaps}
    \vspace{-4mm}
\end{figure}

\begin{figure}[t]
    \centering
    \includegraphics[width=0.8\columnwidth, keepaspectratio]{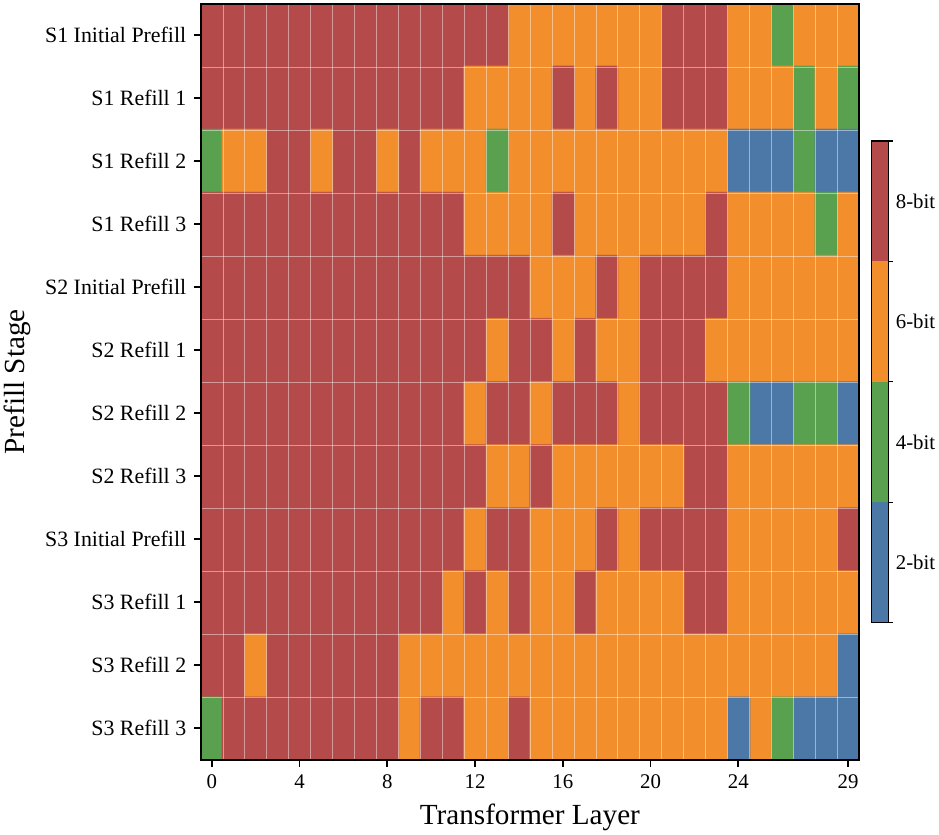}
    \vspace{-5mm}
    \caption{\textbf{Oracle layer-wise precision for the initial prefill and three refills} of Gemma-4-26B on three LongMemEval samples (S1--S3).}
    \label{fig:gemma_lme_refills}
    \vspace{-3mm}
\end{figure}

\section{Additional Motivating Analyses}
\label{sec:appendix_motivation}

\subsection{Layer-Wise Precision Across Decode Steps}
\label{app:layer_decode_precision}

\autoref{fig:layer_decode_heatmaps} shows the oracle per-layer precision at individual decode steps.

\subsection{Layer-Wise Precision Across Prefill and Refills}
\label{app:gemma_lme_refills}

\autoref{fig:gemma_lme_refills} repeats the prefill and refill analysis of \autoref{fig:motivation_1}(b) for an additional model and workload, Gemma-4-26B on LongMemEval, showing the oracle layer-wise precision of the initial prefill and three refills for three samples.

\subsection{Precision Demand by Generation Entropy}
\label{app:entropy_vs_precision}

\begin{figure}[t]
    \centering
    \includegraphics[width=0.7\columnwidth, keepaspectratio]{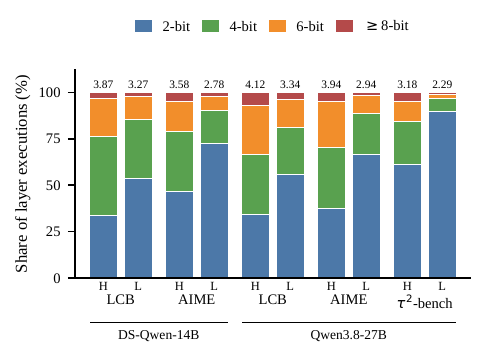}
    \vspace{-5mm}
    \caption{\textbf{Oracle precision demand in high- (H) and low-entropy (L) decode steps.}}
    \vspace{-4mm}
    \label{fig:entropy_vs_precision}
\end{figure}

\autoref{fig:entropy_vs_precision} splits the decode steps of the oracle of \cref{sec:motivation} by $\mathcal{G_E}$ at $\tau = 0.10$. High-entropy steps consistently require higher precision compared to low-entropy steps.

\subsection{Layer-Wise Behavior of $\mathcal{M}_P$ and $\mathcal{M}_D$ Across Models and Datasets}
\label{app:signals_grid}

\begin{figure}[t]
    \centering
    \includegraphics[width=\columnwidth, keepaspectratio]{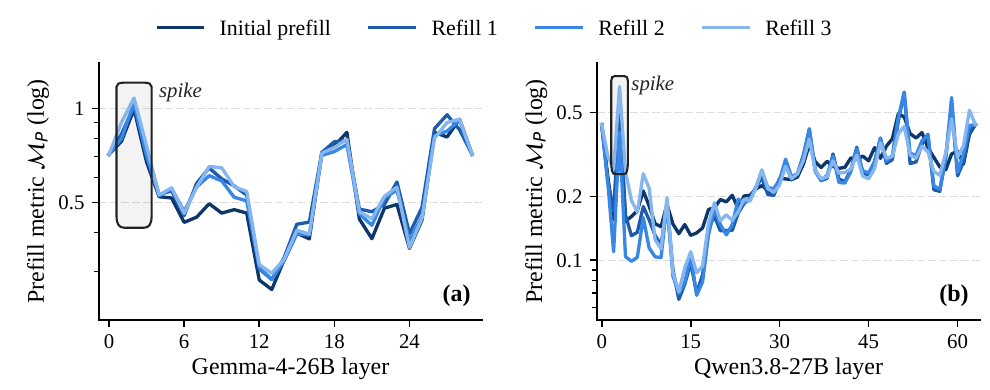}
    % \vspace{-7mm}
    \caption{\textbf{Prefill $\mathcal{M}_P$ for the initial prefill and three refills} of one $\tau^2$-bench task for (a) Gemma-4-26B and (b) Qwen3.8-27B.}
    \label{fig:signals_tau2_refills}
    \vspace{-3mm}
\end{figure}

\begin{figure*}[!t]
    \centering
    \includegraphics[width=0.72\textwidth, keepaspectratio]{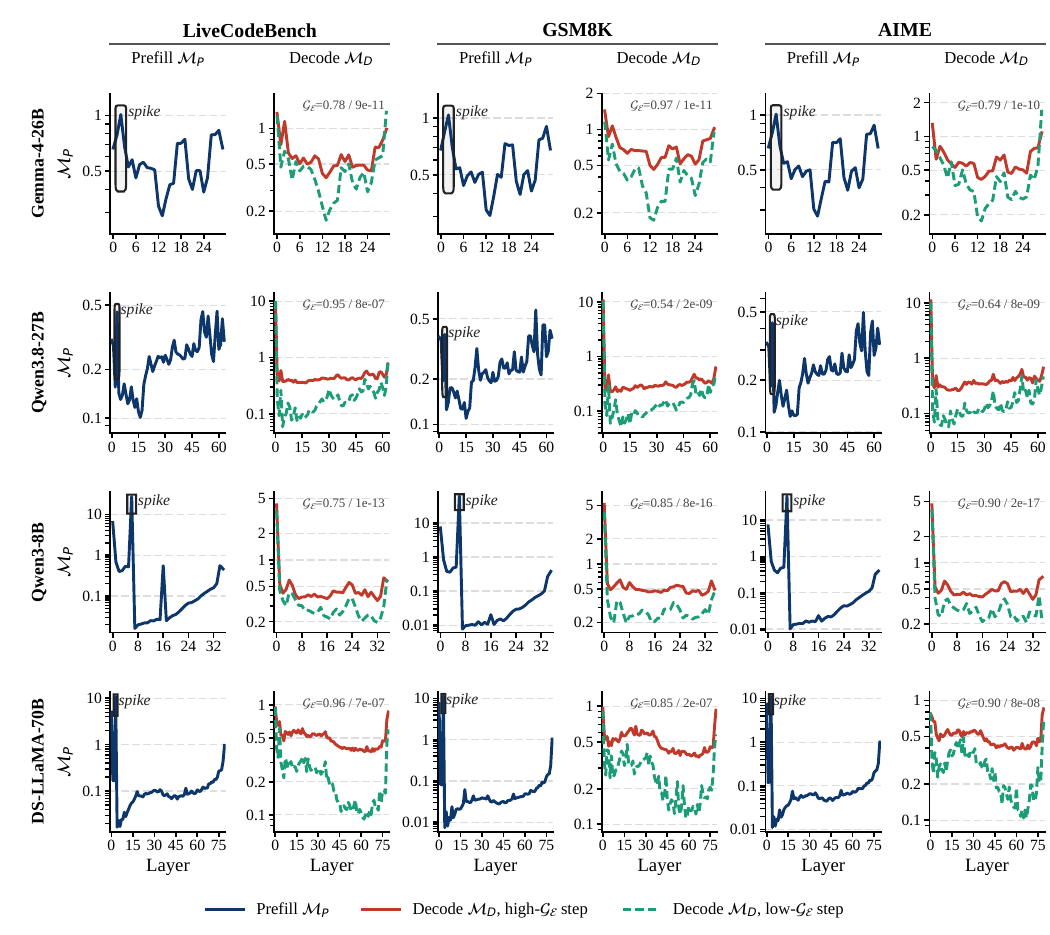}
    \vspace{-6mm}
    \caption{\textbf{Layer-wise $\mathcal{M}_P$ and $\mathcal{M}_D$ across models and datasets.} For each model (rows) and dataset (column pairs), prefill $\mathcal{M}_P$ of one representative sample and decode $\mathcal{M}_D$ of the same sample at the highest- and lowest-$\mathcal{G_E}$ step.}
    \vspace{-3mm}
    % \vspace{-5mm}
    \label{fig:signals_grid}
\end{figure*}

\autoref{fig:signals_grid} extends \autoref{fig:signals} to Gemma-4-26B, Qwen3.8-27B, Qwen3-8B, and DS-LLaMA-70B on LiveCodeBench, GSM8K, and AIME. \autoref{fig:signals_tau2_refills} shows $\mathcal{M}_P$ of the initial prefill and three refills of one $\tau^2$-bench task for Gemma-4-26B and Qwen3.8-27B.

\begin{figure}[t]
    \centering
    \includegraphics[width=\columnwidth, keepaspectratio]{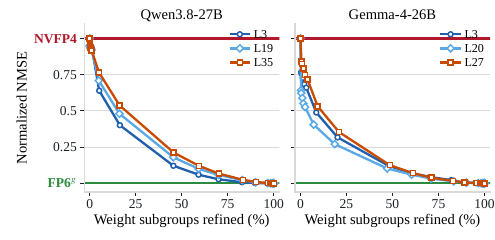}
    \vspace{-5mm}
    \caption{\textbf{Refining a subset of elements achieves NMSE comparable to 6-bit}. Y-axis is NMSE, such that NVFP4 is 1 and FP6$^{\text{g}}$ is 0, vs.\ fraction of subgroups refined to 6-bit, for Qwen3.8-27B and Gemma-4-26B on GSM8K.}
    \label{fig:app_subgroup_models}
    \vspace{-3mm}
\end{figure}

\subsection{Subgroup Refinement on Additional Models}
\label{app:subgroup_models}
\autoref{fig:app_subgroup_models} repeats the experiment of \autoref{fig:motivation_2} on Qwen3.8-27B and Gemma-4-26B with GSM8K, for three layers each.

\begin{figure}[t]
    \centering
    \includegraphics[width=\columnwidth, keepaspectratio]{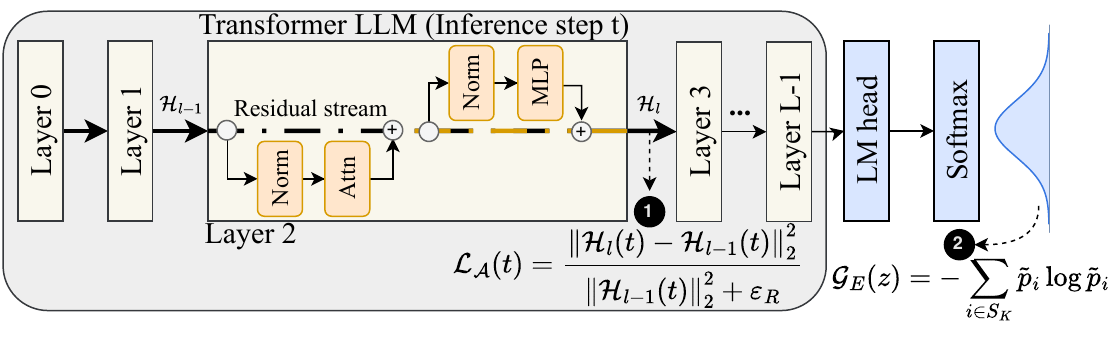}
    \vspace{-3mm}
    \caption{\textbf{Signals used by AutoEMatic for runtime precision adaptation.} \ding{182} Local layer-update activity $\mathcal{L_A}$. \ding{183} Global token-generation uncertainty $\mathcal{G_E}$.}
    \label{fig:methodology}
    \vspace{-3mm}
\end{figure}

\section{Additional AutoEMatic Details}
\label{app:autoematic_details}

\subsection{Signals Used for Runtime Precision Adaptation}
\label{app:signals}

\autoref{fig:methodology} illustrates the two model-internal signals of \cref{sec:runtime_precision_adaptation}.

\subsection{Entropy Signal Across Decode Steps}
\label{app:entropy_steps}

\begin{figure}[t]
    \centering
    \includegraphics[width=\columnwidth, keepaspectratio]{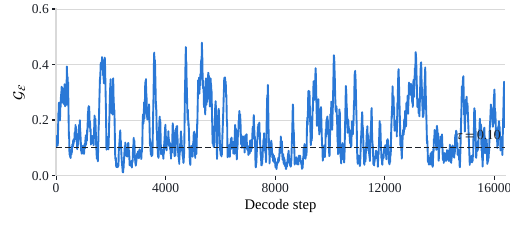}
    \vspace{-3mm}
    \caption{\textbf{Global signal $\mathcal{G_E}$ over a 16K-token AIME generation of Qwen3.8-27B.} The dashed line marks $\tau=0.10$.}
    \label{fig:entropy_steps}
    \vspace{-3mm}
\end{figure}

\autoref{fig:entropy_steps} traces $\mathcal{G_E}$, over a 16{,}383-step AIME generation.

\subsection{Memory Layout}
\label{app:memory_layout}

\begin{figure}[t]
    \centering
    \includegraphics[width=0.75\columnwidth, keepaspectratio]{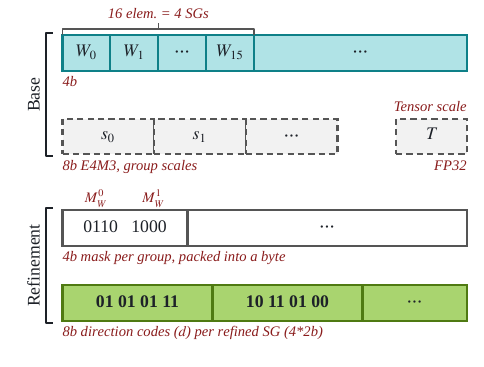}
    \vspace{-5mm}
    \caption{\textbf{AutoEMatic memory layout.}}
    \label{fig:memory_layout}
    \vspace{-3mm}
\end{figure}

\autoref{fig:memory_layout} shows the memory layout of AutoFP, with the NVFP4 base stored separately from the refinement metadata.

\subsection{Extended-$K$ GEMM}
\label{app:extended_k_example}

\begin{figure}[t]
    \centering
    \includegraphics[width=0.9\columnwidth, keepaspectratio]{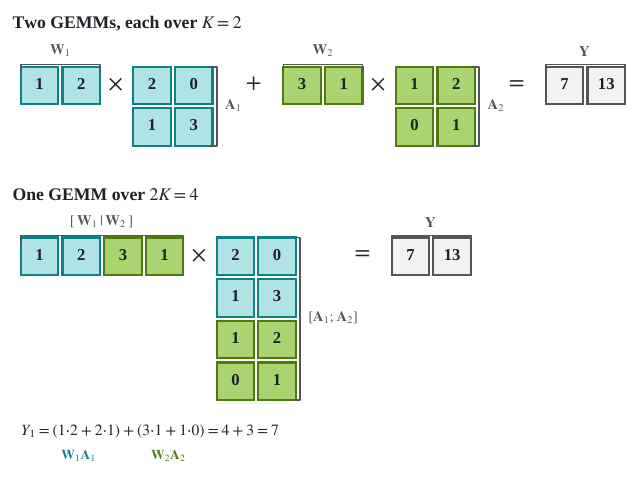}
    \vspace{-5mm}
    \caption{\textbf{A sum of GEMMs as one GEMM over an extended reduction dimension.}}
    \label{fig:extended_k_example}
    \vspace{-3mm}
\end{figure}

\autoref{fig:extended_k_example} illustrates the identity behind the extended-$K$ GEMM of \cref{sec:packed_refinement} on a numeric example: two GEMMs with the same output shape sum to one GEMM over operands concatenated along $K$.

\subsection{Construction of the Extended-$K$ Operands}
\label{app:extended_k_construct}

\begin{figure}[t]
    \centering
    \includegraphics[width=0.95\columnwidth, keepaspectratio]{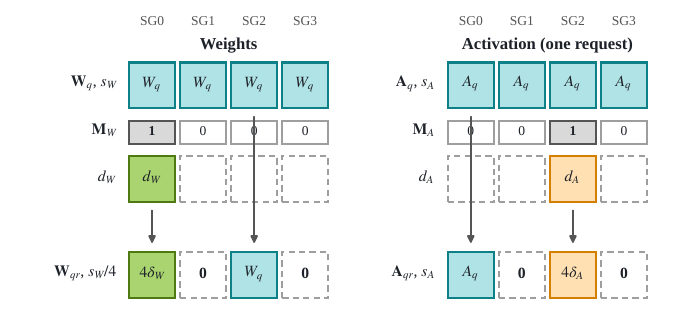}
    \vspace{-3mm}
    \caption{\textbf{Construction of $\mathbf{W}_{qr}$ and $\mathbf{A}_{qr}$ for one 16-element group and one request.} The weight mask places the weight offsets $4\delta_W$ in $\mathbf{W}_{qr}$ and copies the base $A_q$ into $\mathbf{A}_{qr}$; the activation mask copies the base $W_q$ into $\mathbf{W}_{qr}$ and places the activation offsets $4\delta_A$ in $\mathbf{A}_{qr}$; all other positions are zero.}
    \label{fig:extended_k_construct}
    \vspace{-3mm}
\end{figure}

\autoref{fig:extended_k_construct} shows how  $\mathbf{W}_{qr}$ and $\mathbf{A}_{qr}$ are constructed.

\subsection{Numerical Example of the Extended-$K$ GEMM}
\label{app:extended_k_numeric}

\begin{figure}[t]
    \centering
    \includegraphics[width=0.95\columnwidth, keepaspectratio]{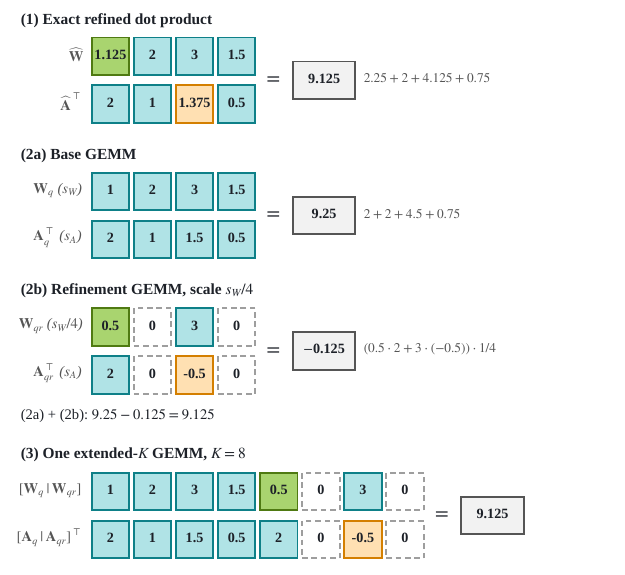}
    \vspace{-3mm}
    \caption{(1)~The exact product of the refined operands $\widehat{\mathbf{W}}$ and $\widehat{\mathbf{A}}$. (2)~The base GEMM on $\mathbf{W}_q$ and $\mathbf{A}_q$ plus a refinement GEMM on the constructed $\mathbf{W}_{qr}$ and $\mathbf{A}_{qr}$ at scale $s_W/4$. (3)~A single GEMM over the concatenated operands.}
    \vspace{-5mm}
    \label{fig:extended_k_numeric}
\end{figure}

\autoref{fig:extended_k_numeric} shows a worked example of the GEMM $[\mathbf{W}_q\,|\,\mathbf{W}_{qr}]\,[\mathbf{A}_q\,|\,\mathbf{A}_{qr}]$:
\begin{equation}
\begin{aligned}
\widehat{\mathbf{W}}\widehat{\mathbf{A}}^{\top} &= \mathbf{W}_q\mathbf{A}_q^{\top} + \tfrac{s_W}{4}\,\mathbf{W}_{qr}\mathbf{A}_{qr}^{\top} \\
&= [\mathbf{W}_q\,|\,\mathbf{W}_{qr}]\,[\mathbf{A}_q\,|\,\mathbf{A}_{qr}]^{\top} \\
&= 9.25 + \tfrac{1}{4}\big(0.5\cdot 2 + 3\cdot(-0.5)\big) = 9.125 .
\end{aligned}
\end{equation}

\subsection{NVFP4-to-AutoFP2 Decomposition}
\label{app:hw-encoding}

The weight encoder decomposes each NVFP4 weight into two AutoFP2 digits, $W_q = c_0 Q_0 + c_1 Q_1$ with $Q_0, Q_1 \in \{-2, \dots, 2\}$ (\cref{sec:hw}). With $c_1 = 1$, to span the maximum NVFP4 magnitude $6$ requires $2c_0 + 2 \ge 6$. $c_0 = 2.5$ satisfies this such that $c_0 Q_0$ provides coarse-grained steps to different NVFP4 binades and $c_1 Q_1$ provides integer corrections $[-2, 2]$, such that their linear combination spans every NVFP4 E2M1 value. Multiplying numerator and denominator by 2 gives integer coefficients such that, $W_q = (5Q_0 + 2Q_1)/2$ and the resulting reconstruction can be realized using only shifts and adds. \Cref{tab:hw-decomposition} lists the digit pair of every positive E2M1 magnitude; negative weights negate both digits.

\begin{table}[t]
\centering
\caption{NVFP4-to-AutoFP2 decomposition.}
\label{tab:hw-decomposition}
\renewcommand*{\arraystretch}{1.05}
\setlength\tabcolsep{5pt}
\begin{tabular}{c|cccccccc}
\Xhline{2\arrayrulewidth}
\rowcolor[HTML]{D3D3D3} $W_q$ & $0$ & $0.5$ & $1$ & $1.5$ & $2$ & $3$ & $4$ & $6$\\
\Xhline{2\arrayrulewidth}
$Q_0$ & $0$ & $1$ & $0$ & $1$ & $0$ & $2$ & $2$ & $2$\\
$Q_1$ & $0$ & $-2$ & $1$ & $-1$ & $2$ & $-2$ & $-1$ & $1$\\
\Xhline{2\arrayrulewidth}
\end{tabular}
\end{table}

For each digit $Q_i$, the encoder produces three bits from the E2M1 bits $\{s, e_1, e_0, m\}$ of the weight: a valid bit $v_i$, which is $1$ if $Q_i \neq 0$ and $0$ if $Q_i = 0$; a sign bit $s_i$; and an exponent bit $e_{q,i}$ with $|Q_i| = 2^{e_{q,i}}$, so that $\{s_i, e_{q,i}\}$ is the E1M0 code of a nonzero digit:
\begin{equation}
\begin{aligned}
 v_0 &= m \lor (e_1 \land e_0), & v_1 &= e_1 \lor e_0 \lor m,\\
 e_{q,0} &= e_1 \land (e_0 \lor m), & e_{q,1} &= \neg e_0 \land (e_1 \lor m),\\
 s_0 &= s, & s_1 &= s \oplus m \oplus (e_1 \land e_0),
\end{aligned}
\label{eq:hw-boolean-decomposition}
\end{equation}
where $\lor$, $\land$, $\neg$, and $\oplus$ denote OR, AND, NOT, and XOR. The valid bits then select the AutoFP2 encoding of \cref{sec:quantization_format}: if both digits are nonzero ($v_0 = v_1 = 1$), the escape bit is $E = 0$ and the fields $F_1$ and $F_0$ hold the two E1M0 codes; otherwise $E = 1$, $F_1$ records which digits are zero, and $F_0$ holds the E1M0 code of the nonzero digit, if any. The valid bits are consumed by this packing and are not stored.

\subsection{Fixed-Point Arithmetic of the vMAC}
\label{app:vmac_fixed_point}
All computation in the vMAC (\cref{sec:vmac}) is fixed-point. Each input activation enters the lanes as the integer $A_k = 2A_q(k)$, since doubling an E2M1 value always yields an integer~\cite{hayashi2026dgemm}. AutoFP2 digits are already integers, whereas in NVFP4 mode the packed digit pair represents twice the NVFP4 weight, $2W_q = 5Q_0 + 2Q_1$ (\cref{sec:weight_encoder}). In the multiply stage each sub-lane forms $A_k F_{j,k}$, $j \in \{0, 1\}$, and within each block the even and odd sub-lanes feed separate adder trees, $A = \sum_{k=0}^{15} A_k F_{0,k}$ and $B = \sum_{k=0}^{15} A_k F_{1,k}$ for lanes 0--15, and $C$, $D$ for lanes 16--31.

In NVFP4 mode, $A$ and $B$ are partial sums over the two digits of the same weights $W_{q,k}$, so by distributivity the mode-dependent reduction unit evaluates $P_0 = 5A + 2B = \sum_k A_k (5F_{0,k} + 2F_{1,k}) = \sum_k (2A_q)(2W_q)$, which is four times the dot product of block 0 in fixed-point; the division by two in $(5A + 2B)/2$ of \cref{sec:vmac} is deferred to the scaling stage. Similarly for $P_1$. In AutoFP2 mode, the even and odd sub-lanes of a block hold weights from two different rows, so the partial sums are summed across blocks as $P_0 = 2(A + C)$ and $P_1 = 2(B + D)$, two independent 32-element dot products; the factor of 2 keeps them at four times the dot product as well.

Each refinement lane forms $R = 2X_q d$ in fixed-point. Since $\delta = d \cdot 2^{-3}$, the floating-point equivalent is $X_q \delta = R/16$, so $R$ is sixteen times the product. The lane contributions are accumulated as $\mathcal{R} = \sum R$, and the base sums $P_j$ are shifted left by two bits to align with them, giving $O_j = 4P_j + \mathcal{R}$ at sixteen times the dot product. For example, with $A_q = 1.5$, $W_q = 3$, and $d_W = -1$: $P = (2 \cdot 1.5)(2 \cdot 3) = 18$, $R = -3$, and $O = 69 = 16 \cdot 1.5\,(3 - 1/8)$. In the scaling stage, each $O_j$ is multiplied by the scale factor of its block and divided by 16, which recovers the floating-point partial sum.

\subsection{Accelerator Floorplan}
\label{app:floorplan}

\begin{figure}[t]
    \centering
    \includegraphics[width=0.6\columnwidth, keepaspectratio]{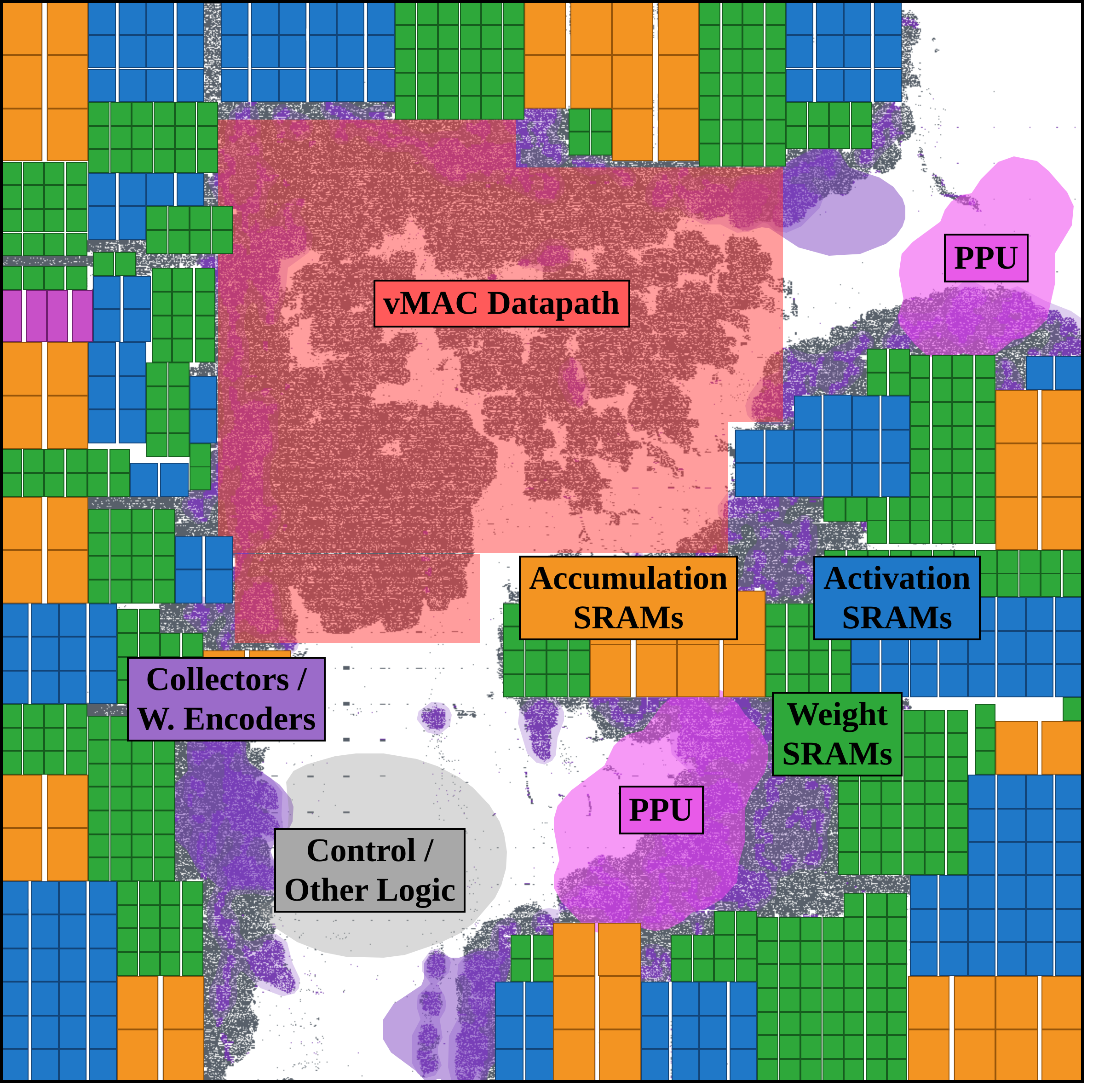}
    \vspace{-3mm}
    \caption{\textbf{Post-layout floorplan of AutoEMatic}.}
    \vspace{-8mm}
    \label{fig:floorplan}
\end{figure}

\autoref{fig:floorplan} shows the AutoEMatic accelerator floorplan.

\section{Refinement-Aware GPTQ Calibration}
\label{app:gptq_calibration}

\Cref{alg:gptq} lists AutoEMatic's offline weight calibration, which extends GPTQ \cite{frantar2023gptq} to an NVFP4 base with $N{:}4$ structured sparse refinement. We follow the NVFP4 notation of \cref{app:nvfp4}: $T$ is the FP32 tensor scale, and $B_{n,j}$ is the E4M3 group scale of row $n$ in group $j$, whose 16 columns we denote $J$. The calibration stores E2M1 codes $q_{n,i}$ and, on refined subgroups, 2-bit signed offsets $d_{n,i}$, decoded with the fixed step of \cref{sec:quantization_format} as $\delta_{n,i} = d_{n,i}\, 2^{-3} \in \mathcal{D} = \{-\tfrac{1}{4}, -\tfrac{1}{8}, 0, +\tfrac{1}{8}\}$; the dequantized weight is $T B_{n,j}\, q_{n,i}$ without refinement (NVFP4) and $T B_{n,j} (q_{n,i} + \delta_{n,i})$ with it (AutoFP6). A refinement is set to $0$ if it would move the element farther from the original weight.

Each group has four 4-element subgroups, and $\mathcal{M}_\ell$ denotes the masks that refine $\ell$ of them; the weight mask $\mathbf{M}_W$ of \cref{sec:quantization_format} holds one such mask $m$ per group. For a mask $m$, $\widehat{\mathbf{W}}(m)$ is the dequantized group that uses the refined values $\widehat{\mathbf{W}}^{r}$ on the selected subgroups and the NVFP4 values $\widehat{\mathbf{W}}^{4}$ elsewhere. Its diagonal-Hessian-weighted error on group $j$ is
\begin{equation}
  \mathcal{E}_j(m) = \sum_{i \in J} \mathbf{H}_{ii} \big\lVert \mathbf{W}_{:,i} - \widehat{\mathbf{W}}_{:,i}(m) \big\rVert_2^2 .
  \label{eq:gptq-group-error}
\end{equation}

The algorithm differs from GPTQ in three ways. \textbf{(1) Budget allocation.} Before the GPTQ pass, a greedy allocation (lines 5--12) distributes $\rho K/4$ refined subgroups across groups, ranked by their reduction in $\mathcal{E}_j$ on the original weights. Each group's mask, which refines $N_W$ of its four subgroups, is shared across all output rows; the remaining $N_A = 4 - N_W$ subgroups are left for activation refinement at runtime. \textbf{(2) Group-wise quantization.} The 16-element group, rather than a single column, is the unit of quantization. Its group scales $B_{n,j}$ are computed from the error-compensated weights when the group is reached, while the tensor scale $T$ stays fixed. \textbf{(3) Group-wise error propagation.} The quantization error of a whole group is propagated to the later columns at once (lines 22--23), rather than one column at a time as in GPTQ.

\begin{algorithm}[t]
\caption{Refinement-aware GPTQ calibration of $\mathbf{W} \in \mathbb{R}^{N \times K}$, given calibration inputs $\mathbf{X} \in \mathbb{R}^{M \times K}$, group size $G = 16$, refinement budget $\rho$, and damping $\omega$. Returns the E2M1 codes $q_{n,i}$, the refinement terms $\delta_{n,i}$ (stored as 2-bit signed offsets) on refined subgroups, the scales $T$ and $B_{n,j}$, and the masks $m^\star_j$.}
\label{alg:gptq}
\renewcommand{\algorithmiccomment}[1]{\hfill{\color{blue}$\triangleright$ #1}}
\newcommand{\LINECOMMENT}[1]{\item[]{\color{blue}$\triangleright$ #1}}
\begin{algorithmic}[1]
\STATE $\mathbf{H} \gets \mathbf{X}^\top \mathbf{X} / M$
\STATE $\mathbf{H}_\omega \gets \mathbf{H} + \omega \operatorname{mean}(\operatorname{diag} \mathbf{H})\, \mathbf{I}$
\STATE $\mathbf{U} \gets \operatorname{Cholesky}(\mathbf{H}_\omega^{-1})^\top$
\STATE $T \gets \operatorname{amax}(\mathbf{W}) / (6 \cdot 448)$ \COMMENT{fixed}
\LINECOMMENT{Allocate refinement budget on the original weights}
\FOR{$j = 0$ \textbf{to} $K/G - 1$}
  \STATE $L_j(\ell) \gets \min_{m \in \mathcal{M}_\ell} \mathcal{E}_j(m)$ for $\ell \in \{0, 1, 2\}$
\ENDFOR
\STATE $\ell_j \gets 0$ for all $j$
\FOR{$b = 1$ \textbf{to} $\rho K / 4$}
  \STATE $j^\star \gets \arg\max_{j : \ell_j < 2} \; L_j(\ell_j) - L_j(\ell_j + 1)$
  \STATE $\ell_{j^\star} \gets \ell_{j^\star} + 1$
\ENDFOR
\LINECOMMENT{GPTQ pass over groups}
\FOR{$j = 0$ \textbf{to} $K/G - 1$}
  \STATE $J \gets jG : (j+1)G$
  \STATE $B_{n,j} \gets Q_{\mathrm{E4M3}}\big(\operatorname{amax}(\mathbf{W}_{n,J}) / 6T\big)$ for all $n$
  \STATE $q_{n,i} \gets Q_{\mathrm{E2M1}}\big(\mathbf{W}_{n,i} / T B_{n,j}\big)$ for all $n$, $i \in J$
  \STATE $\delta_{n,i} \gets \arg\min_{\delta \in \mathcal{D}} \big\lvert \mathbf{W}_{n,i} / T B_{n,j} - q_{n,i} - \delta \big\rvert$
  \STATE $\widehat{\mathbf{W}}^{4}_{n,i} \gets T B_{n,j}\, q_{n,i}$ \COMMENT{dequantized}
  \STATE $\widehat{\mathbf{W}}^{r}_{n,i} \gets T B_{n,j}\, (q_{n,i} + \delta_{n,i})$ \COMMENT{dequantized}
  \STATE $m^\star_j \gets \arg\min_{m \in \mathcal{M}_{\ell_j}} \mathcal{E}_j(m)$
  \STATE $\widehat{\mathbf{W}}_{:,J} \gets \widehat{\mathbf{W}}(m^\star_j)$
  \STATE $\mathbf{E} \gets (\mathbf{W}_{:,J} - \widehat{\mathbf{W}}_{:,J})\, \mathbf{U}_{J,J}^{-1}$
  \STATE $\mathbf{W}_{:,(j+1)G:} \gets \mathbf{W}_{:,(j+1)G:} - \mathbf{E}\, \mathbf{U}_{J,(j+1)G:}$
\ENDFOR
\end{algorithmic}
\end{algorithm}

% \section{Extended-$K$ Operand Layout}
% \label{app:extended_k}

% \Cref{tab:extended_k} shows how the extended-$K$ operands $\mathbf{W}_{qr}$ and $\mathbf{A}_{qr}$ of \cref{sec:packed_refinement} are laid out for one 16-element group. $\mathbf{W}_q$ and $\mathbf{A}_q$ hold the NVFP4 base values. $\mathbf{W}_{qr}$ holds $4\delta_W$ in the subgroup selected by the weight mask, $w_q$ in every subgroup that any request refines in the activations, and zero elsewhere, with the group scale divided by four; each request's $\mathbf{A}_{qr}$ holds $a_q$ in the weight-refined subgroup, $4\delta_A$ in the subgroup it refines, and zeros elsewhere, with its scale unchanged. Reading a column confirms the scaling: SG0 yields $4\delta_W \cdot \tfrac{s_W}{4} \cdot a_q s_A = \delta_W a_q\, s_W s_A$, and SG2 yields $w_q \cdot \tfrac{s_W}{4} \cdot 4\delta_A s_A = w_q \delta_A\, s_W s_A$. A request that does not refine the layer writes $\mathbf{A}_{qr} = \mathbf{0}$ and receives the base product.

\section{GPU Implementation Details}
\label{app:implementation}

AutoEMatic is implemented in vLLM~0.15~\cite{kwon2023efficient} as a quantization plugin, with a custom attention backend and CUDA kernels for Blackwell GPUs. 

\section{Accelerator Simulation Details}
\label{sec:accelerator_simulation}
AutoEMatic performance is measured using a scheduling model calibrated against cycle-accurate Synopsys VCS simulation. Performance data is gathered from these measurements, retaining on-chip data-movement and backpressure costs. Off-chip DRAM transfer time is modeled as transferred bytes divided by an assumed sustained bandwidth of 1 TB/s and added to the execution time.

% \section{GPU Memory Accounting}
% \label{app:memory}

% The bubble areas in \autoref{fig:teaser} are computed with one formula for every method. For Qwen3.8-27B at a 16K-token context,
% \begin{equation}
%  \mathrm{Mem} = U + N_q\,\frac{b_W}{8} + S + E_{\mathrm{KV}}\,\frac{b_{\mathrm{KV}}}{8},
%  \label{eq:memory}
% \end{equation}
% where $U = 6.01$\,GB is the BF16 remainder that no method quantizes (embeddings, output head, and normalization weights), $N_q = 24.35$\,B is the number of elements in the two-dimensional layer weights that every method quantizes, $S = 0.15$\,GB is the precision-independent recurrent and convolution state of the gated-delta-network layers, and $E_{\mathrm{KV}} = 0.54$\,B is the number of key and value elements held by the full-attention layers at 16K tokens. $b_W$ and $b_{\mathrm{KV}}$ are the stored bits per element of the weights and of the KV cache, including all metadata; they are the only quantities that differ between methods.

% Appendix accuracy table: DS-Llama-70B, DS-Qwen-32B, and Llama-3.1-8B-Instruct on the reasoning benchmarks.
% Accuracies in %, one decimal; the calibrated baselines carry their bit budget in the Method column, AutoEMatic per cell.
\providecommand{\bits}[1]{{\scriptsize(#1)}}
\begin{table}[t!]
\centering
\caption{\textbf{Accuracy of AutoEMatic and baselines on DS-Llama-70B, DS-Qwen-32B, and Llama-3.1-8B-Instruct.} Cells report accuracy (\%); execution precision appears in parentheses. LB~v2: LongBench~v2, LCB: LiveCodeBench. Llama-3.1-8B-Instruct is not evaluated on AIME and LCB, where its BF16 accuracy is too low for a meaningful comparison.}
\label{tab:main_results_ds70}
\renewcommand*{\arraystretch}{1.05}
\setlength\tabcolsep{2pt}
\resizebox{0.9\columnwidth}{!}{%
\begin{tabular}{ll|cccc}
\Xhline{2\arrayrulewidth}
\rowcolor[HTML]{D3D3D3} \textbf{Model} & \textbf{Method (budget)} & LB~v2 & AIME & LCB & GSM8K \\
\Xhline{2\arrayrulewidth}
\multirow{8}{*}{\rotatebox[origin=c]{90}{DS-Llama-70B}}
 & \cellcolor[HTML]{EBEBEB}BF16 & \cellcolor[HTML]{EBEBEB}41.0 & \cellcolor[HTML]{EBEBEB}66.7 & \cellcolor[HTML]{EBEBEB}68.2 & \cellcolor[HTML]{EBEBEB}94.5 \\ \cline{2-6}
 & NVFP4 & 29.8 & 53.3 & 51.7 & 90.6 \\
 & FP6$^{\text{g}}$ & \textbf{41.0} & \textbf{66.7} & \textbf{66.2} & \textbf{94.5} \\
 & AutoQuantize \bits{5.3} & 38.7 & \textbf{66.7} & 55.7 & \textbf{94.5} \\
 & FlexQuant \bits{5.3} & 27.8 & 50.0 & 48.6 & 86.5 \\
 & DP-LLM (weight-only) \bits{5.3} & 21.0 \bits{5.1} & 43.3 \bits{5.3} & 43.3 \bits{5.3} & 81.6 \bits{5.1} \\
 & DP-LLM$^\dagger$ \bits{5.3} & 19.0 \bits{5.1} & 36.7 \bits{5.3} & 39.5 \bits{5.3} & 77.6 \bits{5.1} \\
 & \cellcolor[HTML]{D5E8D4}\textbf{AutoEMatic} & \cellcolor[HTML]{D5E8D4}\textbf{41.0} \bits{5.4} & \cellcolor[HTML]{D5E8D4}\textbf{66.7} \bits{5.1} & \cellcolor[HTML]{D5E8D4}65.6 \bits{4.5} & \cellcolor[HTML]{D5E8D4}\textbf{94.5} \bits{4.9} \\
\Xhline{1\arrayrulewidth}
\multirow{8}{*}{\rotatebox[origin=c]{90}{DS-Qwen-32B}}
 & \cellcolor[HTML]{EBEBEB}BF16 & \cellcolor[HTML]{EBEBEB}49.8 & \cellcolor[HTML]{EBEBEB}73.3 & \cellcolor[HTML]{EBEBEB}54.2 & \cellcolor[HTML]{EBEBEB}92.5 \\ \cline{2-6}
 & NVFP4 & 43.5 & 60.0 & 50.5 & 87.6 \\
 & FP6$^{\text{g}}$ & \textbf{49.8} & \textbf{70.0} & \textbf{52.6} & \textbf{90.5} \\
 & AutoQuantize \bits{4.8} & \textbf{49.8} & 66.7 & 51.2 & 89.3 \\
 & FlexQuant \bits{4.8} & 40.5 & 56.7 & 46.9 & 81.5 \\
 & DP-LLM (weight-only) \bits{4.8} & 36.6 \bits{4.8} & 50.0 \bits{4.8} & 45.0 \bits{4.8} & 76.5 \bits{4.5} \\
 & DP-LLM$^\dagger$ \bits{4.8} & 30.6 \bits{4.8} & 36.7 \bits{4.8} & 40.7 \bits{4.8} & 71.8 \bits{4.5} \\
 & \cellcolor[HTML]{D5E8D4}\textbf{AutoEMatic} & \cellcolor[HTML]{D5E8D4}\textbf{49.8} \bits{4.6} & \cellcolor[HTML]{D5E8D4}\textbf{70.0} \bits{4.8} & \cellcolor[HTML]{D5E8D4}52.4 \bits{4.6} & \cellcolor[HTML]{D5E8D4}\textbf{90.5} \bits{4.4} \\
\Xhline{1\arrayrulewidth}
\multirow{8}{*}{\rotatebox[origin=c]{90}{\shortstack{Llama-3.1-\\8B-Instruct}}}
 & \cellcolor[HTML]{EBEBEB}BF16 & \cellcolor[HTML]{EBEBEB}30.5 & \cellcolor[HTML]{EBEBEB}-- & \cellcolor[HTML]{EBEBEB}-- & \cellcolor[HTML]{EBEBEB}82.8 \\ \cline{2-6}
 & NVFP4 & 27.5 & -- & -- & 75.6 \\
 & FP6$^{\text{g}}$ & \textbf{30.5} & -- & -- & \textbf{80.5} \\
 & AutoQuantize \bits{4.5} & \textbf{30.5} & -- & -- & 78.9 \\
 & FlexQuant \bits{4.5} & 25.9 & -- & -- & 71.6 \\
 & DP-LLM (weight-only) \bits{4.5} & 24.6 \bits{4.4} & -- & -- & 63.8 \bits{4.3} \\
 & DP-LLM$^\dagger$ \bits{4.5} & 22.8 \bits{4.4} & -- & -- & 59.6 \bits{4.3} \\
 & \cellcolor[HTML]{D5E8D4}\textbf{AutoEMatic} & \cellcolor[HTML]{D5E8D4}\textbf{30.5} \bits{4.4} & \cellcolor[HTML]{D5E8D4}-- & \cellcolor[HTML]{D5E8D4}-- & \cellcolor[HTML]{D5E8D4}\textbf{80.5} \bits{4.1} \\
\Xhline{2\arrayrulewidth}
\end{tabular}}
\vspace{-5mm}
\end{table}

\section{Additional Accuracy Results}
\label{app:results}

\autoref{tab:main_results_ds70} reports the accuracy of AutoEMatic and the baselines on DS-Llama-70B, DS-Qwen-32B, and Llama-3.1-8B-Instruct on the reasoning benchmarks.

\section{Additional Ablations}
\label{app:ablations}

\begin{figure*}[t]
    \centering
    \includegraphics[width=0.8\textwidth]{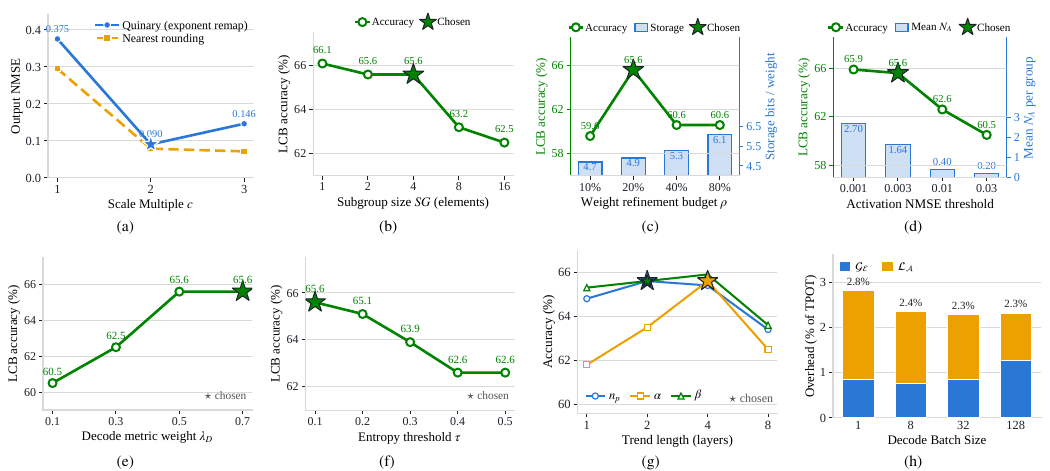}
    \vspace{-3mm}
    \caption{\textbf{Additional ablations} on DS-LLaMA-70B (LCB). (a) Scale multiple $c$ of AutoFP2: mean layer-output NMSE over all 80 layers at W2A4. (b) Subgroup size $SG$ with $20\%$ of all weight subgroups refined. (c) Weight refinement budget $\rho$, with storage precision. (d) Activation NMSE threshold, with the mean number of activation subgroups per group exceeding it. (e) $\lambda_D$ sweep. (f) Entropy threshold $\tau$. (g) Trend lengths $n_p$, $\alpha$, and $\beta$. (h) Overhead of computing $\mathcal{G_E}$ and $\mathcal{L_A}$ as a share of decode TPOT. Stars mark the chosen settings.}
    \vspace{-6mm}
    \label{fig:abl_appendix}
\end{figure*}

\begin{figure}[t]
    \centering
    \includegraphics[width=0.7\columnwidth, keepaspectratio]{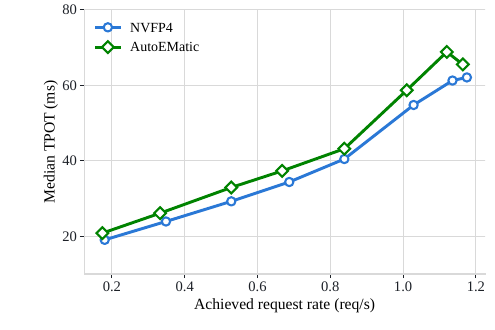}
    \vspace{-3mm}
    \caption{\textbf{Online serving under continuous batching.} Median TPOT vs.\ achieved request rate for DS-LLaMA-70B on $4\times$ GB300 (TP=4) with Poisson arrivals of LongBench~v2 requests.}
    \label{fig:online_serving}
    \vspace{-3mm}
\end{figure}

\noindent
\textbf{AutoFP2 scale multiple $c$. } \autoref{fig:abl_appendix}(a) sweeps the scale multiple $c$ of \cref{app:scale-sharing}: $c = 2$ gives an NMSE of $0.09$, within $1.15\times$ of nearest rounding, whereas $c = 1$ clips the large magnitudes and $c = 3$ places the merged E2M1 codes $\{1, 1.5, 2, 3\}$ at $3$, raising the NMSE to $0.38$ and $0.15$.

\begin{figure}[t]
    \centering
    \includegraphics[width=0.6\columnwidth, keepaspectratio]{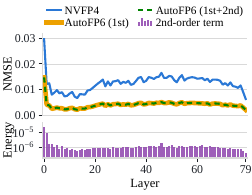}
    \vspace{-3mm}
    \caption{\textbf{Second-order refinement term.} Per-layer output NMSE of AutoFP6 with and without the second-order product $\mathbf{W}_r\mathbf{A}_r$ (top) and the energy of that product relative to the layer output (bottom) on DS-LLaMA-70B.}
    \label{fig:abl_second_order}
    \vspace{-3mm}
\end{figure}

\noindent
\textbf{Impact of second-order refinement. }\label{app:abl_second_order}\autoref{fig:abl_second_order} measures the second-order refinement product $\mathbf{W}_r\mathbf{A}_r$ that the extended-$K$ GEMM omits (\cref{sec:packed_refinement}). Its energy is $1.4\times10^{-6}$ of the per-layer output, and consequently the AutoFP6 output NMSE is $0.0041$ with and without this term.

% \begin{figure}[t]
%     \centering
%     \includegraphics[width=0.9\columnwidth, keepaspectratio]{figures/autofp2_storage.pdf}
%     \vspace{-3mm}
%     \caption{\textbf{Storing AutoFP2 instead of NVFP4.} Latency of a Qwen3-8B decoder layer on a GB300 when weights are stored as packed AutoFP2 digit pairs and NVFP4 is reconstructed at runtime, normalized to storing NVFP4 directly.}
%     \label{fig:autofp2_storage}
% \end{figure}
%
% \noindent
% \textbf{AutoFP2 as the stored format. }\label{app:autofp2_storage}NVFP4 cannot be recovered from a single AutoFP2 digit; the only AutoFP2 form from which it is recoverable is the digit pair of \cref{sec:weight_encoder}, which takes 5 bits per weight. \autoref{fig:autofp2_storage} measures the cost of storing these pairs and reconstructing NVFP4 before each GEMM: the layer latency is $1.56\times$ that of storing NVFP4 directly at batch size 1 and $1.15\times$ at batch size 4096, since the reconstruction adds a pass over the weights that is exposed when the GEMM is bandwidth-bound. NVFP4 is therefore the stored base and AutoFP2 is derived from it on demand (\cref{sec:quantization_format}).
%
% \begin{itemize}[leftmargin=*, topsep=2pt, itemsep=0pt]
%     \item \tb{TODO: write the experiment setup for \autoref{fig:autofp2_storage} (model, layers measured, how NVFP4 is reconstructed from the digit pairs, timing method).}
% \end{itemize}

\begin{figure}[t]
    \centering
    \includegraphics[width=0.85\columnwidth, keepaspectratio]{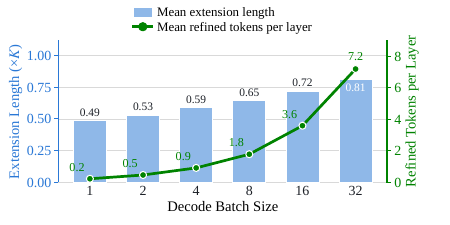}
    \vspace{-3mm}
    \caption{\textbf{Extension of the reduction dimension} of refined GEMMs on Qwen3.8-27B $\tau^2$-bench vs.\ decode batch size, with the mean number of refined tokens per layer.}
    \label{fig:abl_extended_k_length}
    \vspace{-3mm}
\end{figure}

\noindent
\textbf{Extension of $K$. }\label{app:abl_extended_k_length}The reduction extension of a refined GEMM depends on the \# of subgroups refined. On Qwen3.8-27B $\tau^2$-bench (\autoref{fig:abl_extended_k_length}), the extension grows from $0.49K$ at batch 1 to $0.81K$ at batch 32 with average refined requests per layer rising from $0.2$ to $7.2$.

\begin{table}[t]
\centering
\caption{\textbf{Extension to 8 bits} with DS-LLaMA-70B on LCB; rows 1 and 7 of \autoref{tab:ablation_components} are repeated for reference.}
\label{tab:ablation_autofp8}
\renewcommand*{\arraystretch}{1.05}
\setlength\tabcolsep{3pt}
\resizebox{\columnwidth}{!}{%
\begin{tabular}{l|c|c|c}
\Xhline{2\arrayrulewidth}
\rowcolor[HTML]{D3D3D3} \textbf{Component} & \textbf{Bits} & \textbf{Acc.} & \textbf{2 / 4 / 6 / 8 (\%)} \\
\Xhline{2\arrayrulewidth}
\cellcolor[HTML]{EBEBEB}BF16 & \cellcolor[HTML]{EBEBEB}16 & \cellcolor[HTML]{EBEBEB}68.4 & \cellcolor[HTML]{EBEBEB}-- \\
AutoEMatic format + runtime policy, $\mathcal{P}=\{2,4,6\}$ & 4.9 & 65.6 & 15 / 45 / 40 / 0 \\
\hline
\cellcolor[HTML]{D5E8D4}AutoFP8 + Refinement-aware GPTQ & \cellcolor[HTML]{D5E8D4}5.3 & \cellcolor[HTML]{D5E8D4}67.9 & \cellcolor[HTML]{D5E8D4}0 / 0 / 0 / 100 \\
\cellcolor[HTML]{D5E8D4}\textbf{AutoEMatic format + runtime policy}, $\mathcal{P}=\{2,4,6,8\}$ & \cellcolor[HTML]{D5E8D4}5.3 & \cellcolor[HTML]{D5E8D4}\textbf{67.9} & \cellcolor[HTML]{D5E8D4}30 / 35 / 15 / 20 \\
\Xhline{2\arrayrulewidth}
\end{tabular}}
\end{table}

\noindent
\textbf{Extension to 8 bits. }\label{app:abl_autofp8}Widening the signed offsets from 2 to 4 bits extends AutoEMatic to AutoFP8 (\autoref{tab:ablation_autofp8}). With every layer statically refined to AutoFP8 and refinement-aware GPTQ, DS-LLaMA-70B attains $67.9\%$ on LCB at $5.3$ stored bits per weight. The runtime policy with $\mathcal{P}=\{2,4,6,8\}$ attains the same $67.9\%$, within $0.5\%$ of BF16, at $5.3$ bits while executing $30\%$ of layers at 2-bit, $35\%$ at 4-bit, $15\%$ at 6-bit, and $20\%$ at 8-bit.

\noindent
\textbf{Subgroup size. }\label{app:abl_subgroup_size}\autoref{fig:abl_appendix}(b) varies $SG$ with $20\%$ of all weight subgroups refined. Finer subgroups raise accuracy only from $65.6\%$ at $SG=4$ to $66.1\%$ at $SG=1$, however, finer subgroups require more metadata per group and are not as performant. Coarser subgroups lose $2.4$--$3.1\%$, as a refined subgroup then spends signed offsets on elements that do not need them.

\noindent
\textbf{Weight refinement budget. }\label{app:abl_rho}\autoref{fig:abl_appendix}(c) sweeps $\rho$, the fraction of all weight subgroups refined, over $\{10, 20, 40, 80\}\%$, with storage of $4.5 + 2\rho$ bits per weight. Accuracy peaks at $\rho = 20\%$ ($65.6\%$ at $4.9$ bits) and falls on both sides: at $10\%$ too few weight subgroups are refined, while at $40\%$ and $80\%$ the weight mask occupies subgroups that the activations would otherwise refine at runtime ($N_A = 4 - N_W$), so additional weight bits displace activation refinement and accuracy drops to $60.6\%$.

\noindent
\textbf{Activation refinement threshold. }\label{app:abl_nmse}\autoref{fig:abl_appendix}(d) sweeps the NMSE threshold above which an activation subgroup is refined, over $\{0.001, 0.003, 0.01, 0.03\}$, together with the mean number of subgroups per group that exceed it. Lowering the threshold below $0.003$ refines $2.7$ rather than $1.6$ subgroups per group for $0.3\%$ more accuracy, while raising it to $0.01$ and $0.03$ leaves most groups unrefined.

\noindent
\textbf{$\mathbf{\lambda_D}$. }\label{app:abl_lambda}\autoref{fig:abl_appendix}(e) sweeps $\lambda_D$ over $\{0.1, 0.3, 0.5, 0.7\}$.

\noindent
\textbf{Entropy threshold. }\label{app:abl_tau}\autoref{fig:abl_appendix}(f) sweeps $\tau$, which separates the high- and low-entropy decode regimes, over $\{0.1, \dots, 0.5\}$. Accuracy is highest at $\tau = 0.1$ ($65.6\%$) and declines monotonically to $62.6\%$ at $\tau \ge 0.4$, as more steps with several plausible continuations are treated as low-entropy and allowed to lower precision aggressively.

\noindent
\textbf{Trend lengths. }\label{app:trend_lengths}\autoref{fig:abl_appendix}(g) sweeps $n_p$, $\alpha$, and $\beta$ over $\{1,2,4,8\}$.

% \noindent
% \textbf{Spike replay. }\label{app:spike_replay}During decode, replaying a layer that spikes while executing below 4 bits (\cref{sec:runtime_precision_adaptation}) reduces TPOT by less than $3\%$ on the steps in which a spike occurs, and spikes occur on $1$--$2\%$ of decode steps on average across the evaluated models.

\noindent
\textbf{Signal overhead. }\label{app:signal_overhead}\autoref{fig:abl_appendix}(h) reports the cost of computing the signals as a share of decode TPOT. $\mathcal{G_E}$ is one top-$K$ selection and entropy over the logits per step, below $1.3\%$ at every batch size; $\mathcal{L_A}$ is a per-layer reduction of the residual update, $2.4\,\mu$s per layer, whose share falls from $2.0\%$ at batch 1 to $1.1\%$ at batch 128 as it amortizes over rows. Together the signals add $2.3$--$2.8\%$ of TPOT.

\noindent
\textbf{Online serving. }\label{app:online_serving}\autoref{fig:online_serving} serves DS-LLaMA-70B in vLLM under continuous batching with Poisson request arrivals at increasing rates, using LongBench~v2 prompts of about 8K tokens with 1--2K output tokens.

\begin{figure}[t]
    \centering
    \includegraphics[width=0.8\columnwidth, keepaspectratio]{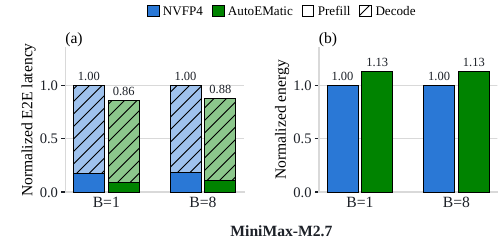}
    \vspace{-5mm}
    \caption{\textbf{Accelerator latency and energy at iso-accuracy with NVFP4.}}
    \vspace{-5mm}
    \label{fig:accel_iso_nvfp4}
\end{figure}

% Instantaneous vs trend-based precision changes, Qwen3-8B on GSM8K (200 problems, 1024-token cap, thinking off).
% Bits = mean nominal precision level over executed layers. Switches/100 = precision changes between adjacent layers per 100 decode layer executions.
\begin{table}[!htb]
\centering
\caption{\textbf{Instantaneous vs.\ trend-based precision changes} with Qwen3-8B on GSM8K.}
\label{tab:instant_vs_trend}
\renewcommand*{\arraystretch}{1.05}
\setlength\tabcolsep{3pt}
\resizebox{\columnwidth}{!}{%
\begin{tabular}{l|c|c|c}
\Xhline{2\arrayrulewidth}
\rowcolor[HTML]{D3D3D3} \textbf{Policy} & \textbf{Bits} & \textbf{Acc.} & \textbf{2 / 4 / 6 (\%)} \\
\Xhline{2\arrayrulewidth}
Instantaneous ($n_p{=}\alpha{=}\beta{=}1$) & 4.7 & 92.5 & 0 / 66 / 34 \\
\hline
\cellcolor[HTML]{D5E8D4}\textbf{AutoEMatic} ($n_p{=}2$, $\alpha{=}4$, $\beta{=}2$) & \cellcolor[HTML]{D5E8D4}4.6 & \cellcolor[HTML]{D5E8D4}\textbf{95.5} & \cellcolor[HTML]{D5E8D4}0 / 69 / 31 \\
\Xhline{2\arrayrulewidth}
\end{tabular}}
\end{table}

\noindent
\textbf{Accelerator latency and energy at iso-accuracy with NVFP4. }\label{app:accel_iso_nvfp4}The main comparison (\cref{sec:hw}) holds accuracy at AutoEMatic's level. \autoref{fig:accel_iso_nvfp4} instead holds it at NVFP4's. This enables the NVFP4 datapath to run all layers at its native precision. To achieve iso-accuracy with NVFP4, we tune AutoEMatic assignments on MiniMax-M2.7 such that AutoEMatic assigns $30\%$ of layers to AutoFP2, $60\%$ to NVFP4, and $10\%$ to AutoFP6 and matches the accuracy of the NVFP4. As shown in \autoref{fig:accel_iso_nvfp4}, AutoEMatic still reduces end-to-end latency by $12$--$14\%$, with prefill $1.7$--$2.0\times$ faster at batch sizes 1 and 8, at $13\%$ higher energy, due to the additional components in the mixed-precision AutoEMatic datapath.

\begin{figure}[t!]
    \centering
    \includegraphics[width=0.65\columnwidth, keepaspectratio]{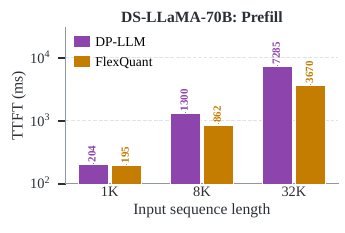}
    \vspace{-5mm}
    \caption{\textbf{TTFT of the runtime-adaptive baselines} DP-LLM and FlexQuant on DS-LLaMA-70B vs.\ input sequence length.}
    \vspace{-5mm}
    \label{fig:baseline_ttft}
\end{figure}

\noindent
\textbf{DP-LLM and FlexQuant performance. }\label{app:baseline_perf}\autoref{fig:baseline_ttft} reports the prefill latency of DP-LLM and FlexQuant on DS-LLaMA-70B. At batch size 1 they decode at $5.9$ and $7.3$ tokens/s, respectively, against $131$ tokens/s for NVFP4 and $120$ for AutoEMatic in \autoref{fig:gpu_perf}.

% \begin{figure}[t]
%     \centering
%     \includegraphics[width=\columnwidth, keepaspectratio]{figures/abl_instant_vs_trend.pdf}
%     \caption{\textbf{Instantaneous vs.\ trend-based precision changes.} Per-layer precision over 60 consecutive decode steps of one GSM8K request on Qwen3-8B under instantaneous decisions (left) and the trend-based policy (right).}
%     \label{fig:instant_vs_trend}
% \end{figure}

\noindent
\textbf{Instantaneous vs.\ trend-based changes. }\label{app:instant_vs_trend}\autoref{tab:instant_vs_trend} compares making precision decisions instantaneously versus AutoEMatic's trend-based policy. The trend-based policy reaches both lower average execution precision ($4.6$ vs.\ $4.7$ bits) and higher accuracy ($95.5\%$ vs.\ $92.5\%$). Instantaneous decisions react to layer-to-layer noise in the metric, the trend rule acts only on sustained changes, which filters this noise.

\section{Limitations}
As it stands, while AutoFP supports the precision set $\mathcal{P}=\{2,4,6\}$, it cannot take advantage of AutoFP2's memory bandwidth benefits. Using AutoFP2 as the stored base would require reconstructing NVFP4 for most layer executions, introducing overhead that can outweigh the bandwidth savings during decode. Furthermore, AutoFP2's coarse representation requires larger corrections to recover NVFP4 values, which the current signed-offset encoding does not directly support. Future work will work on remedying this.

\section{AI Usage Statement}
Portions of this paper were refined with the assistance of a large language model (LLM), specifically OpenAI GPT 6-Sol, used to polish writing and help reduce verbosity to meet page limit. The LLM also assisted in writing the python code for making several figures for the data the authors independently collected. All technical content, methods, and results were conceived entirely by the authors, without influence from any AI tool. The AI was however used to run parameter sweeps, conduct design space exploration, suggest optimizations (which the authors thoroughly verified) and also launch and monitor accuracy and performance experiments.

%%%%%%%%%%%%%%%%%%%%%%%%%%%%%%%%%%%%%%%%%%%%%%%%%%%%%%%%%%%%%%%%%%%%%%%%%%%%%%%
%%%%%%%%%%%%%%%%%%%%%%%%%%%%%%%%%%%%%%%%%%%%%%%%%%%%%%%%%%%%%%%%%%%%%%%%%%%%%%%

\end{document}